%% file: main.tex
\documentclass[10pt,twocolumn,letterpaper]{article}

\usepackage{cvpr}              

\input{preamble}

\definecolor{cvprblue}{rgb}{0.21,0.49,0.74}
\usepackage[pagebackref,breaklinks,colorlinks,citecolor=cvprblue]{hyperref}

\usepackage{booktabs}
\usepackage{color}
\usepackage{bbding} 
\usepackage{amsfonts} 
\usepackage{algorithm}
\usepackage{algpseudocode}
\usepackage{multirow}
\usepackage{stfloats}

\definecolor{cred}{HTML}{FF6B6B}
\definecolor{cyellow}{HTML}{FEC260}
\definecolor{cgreen}{HTML}{70AD47}
\definecolor{cblue}{HTML}{4D96FF}
\definecolor{cpurple}{HTML}{2A0944}
\definecolor{ggray}{RGB}{127,127,127}
\definecolor{aliceblue}{rgb}{0.94, 0.97, 1.0}

\usepackage{graphicx}
\usepackage{caption}

\def\paperID{4256} 
\def\confName{CVPR}
\def\confYear{2024}

\title{FreestyleRet: Retrieving Images from Style-Diversified Queries}

\author{
Hao Li$^{1,2}$\footnotemark[1], Curise Jia$^{1}$\footnotemark[1], Peng Jin$^{1}$, Zesen Cheng$^{1}$, Kehan Li$^{1}$, Jialu Sui$^{4}$, Chang Liu$^{3}$, Li Yuan$^{1,2}$\footnotemark[2]\\
\small{$^1$School of Electronic and Computer Engineering, Peking University Shenzhen Graduate School}, \\ \small{$^2$Peng Cheng Laboratory, Shenzhen}, \small{$^3$Department of Automation and BNRist, Tsinghua University}, \\ \small{$^4$School of
Science and Engineering, Chinese University of Hong Kong, Shenzhen} \\
{\tt\small \{lihao1984, yuanli-ece\}@pku.edu.cn},\quad {\tt\small liuchang2022@tsinghua.edu.cn}
}

\begin{document}
\twocolumn[{%
\renewcommand\twocolumn[1][]{#1}%
\maketitle
\begin{center}
    \centering
    \includegraphics[width=1.\linewidth]{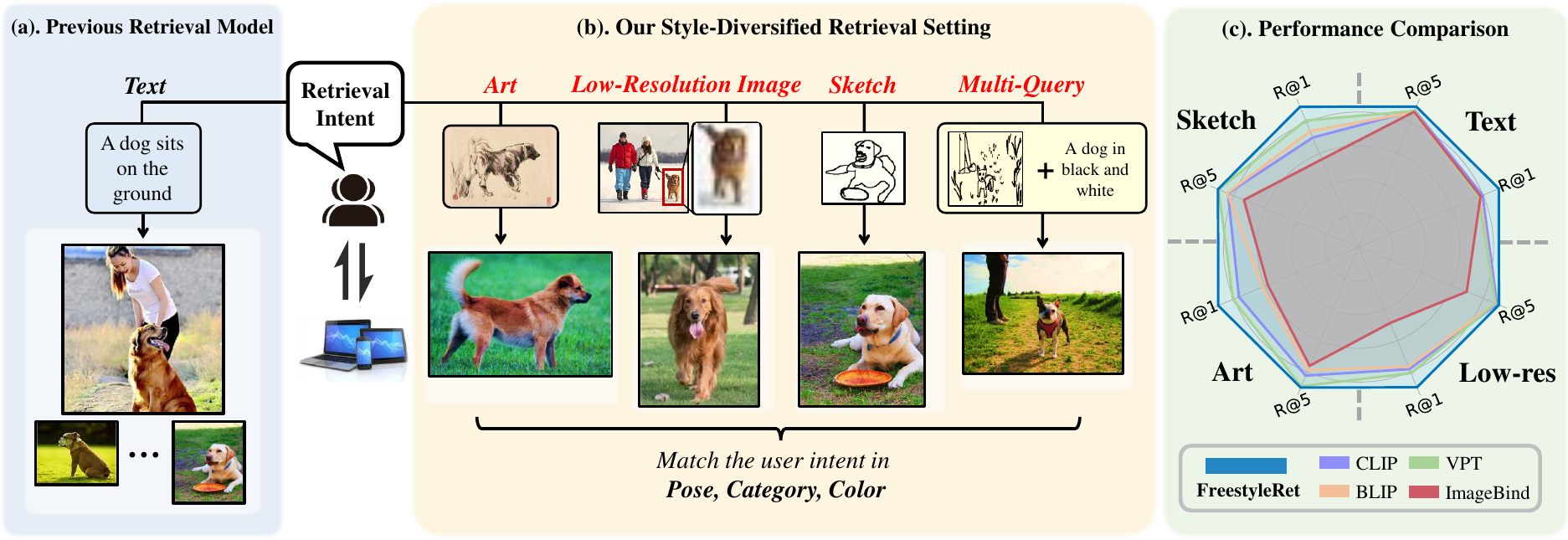}
    \captionof{figure}{
    (a). Previous Retrieval Models focus on text-query retrieval exploration, neglecting the retrieval ability for other query styles. 
    (b). Our style-diversified retrieval setting considers the various query styles that users may prefer, including sketch, art, low-resolution, text, and their combination, including sketch+text, art+text, etc. Our model makes fine-grained retrieval based on the shape, color, and pose features from style-diversified query inputs.
    (c). The performance comparison between our model and other retrieval baselines.
    }
    \label{fig:motivation}
\end{center}%
}]

\input{sec/0_abstract}
\input{sec/1_intro}
\input{sec/2_related}
\input{sec/3_method}
\input{sec/4_exp}
\input{sec/5_conclusion}
\clearpage
\input{sec/6_suppl}
{
    \small
    \bibliographystyle{ieeenat_fullname}
    \bibliography{reference}
}


\end{document}

%% file: preamble.tex
\usepackage[dvipsnames,table]{xcolor}


%% file: sec/0_abstract.tex
\begin{abstract}

Image Retrieval aims to retrieve corresponding images based on a given query. In application scenarios, users intend to express their retrieval intent through various query styles. 
However, current retrieval tasks predominantly focus on text-query retrieval exploration, leading to limited retrieval query options and potential ambiguity or bias in user intention. 
In this paper, we propose the Style-Diversified Query-Based Image Retrieval task, which enables retrieval based on various query styles. To facilitate the novel setting, we propose the first Diverse-Style Retrieval dataset, encompassing diverse query styles including text, sketch, low-resolution, and art.
We also propose a light-weighted style-diversified retrieval framework. For various query style inputs, we apply the Gram Matrix to extract the query's textural features and cluster them into a style space with style-specific bases. Then we employ the style-init prompt tuning module to enable the visual encoder to comprehend the texture and style information of the query.
Experiments demonstrate that our model, employing the style-init prompt tuning strategy, outperforms existing retrieval models on the style-diversified retrieval task. Moreover, style-diversified queries~(sketch+text, art+text, etc) can be simultaneously retrieved in our model. The auxiliary information from other queries enhances the retrieval performance within the respective query\footnote{$*$ Equal Contribution, $\dagger$ Corresponding Author. We have included the code and dataset in the supplementary material.}. The code is available in  \url{https://github.com/CuriseJia/FreeStyleRet}.


\end{abstract}

%% file: sec/1_intro.tex
\section{Introduction}
\label{sec:introduction}

Query-based image retrieval~(QBIR)~\cite{thomee2012interactive} refers to the task of retrieving relevant images from a large image database based on the user's query or search term. QBIR has numerous applications, ranging from image search engines~\cite{isinkaye2015recommendation} to cross-modality downstream tasks~\cite{li2023weakly,li2022joint}. It plays a crucial role in enabling users to locate and obtain related visual content based on their retrieval intent.


The diversification of user retrieval intents poses a significant and unresolved problem in QBIR~\cite{li2021recent}. Selecting appropriate queries to express user intents and enabling models to accommodate diverse query styles are crucial challenges. However, the current exploration in the field of QBIR has primarily focused on text-image retrieval~\cite{radford2021learning,li2022blip} and text-video retrieval~\cite{jin2023diffusionret,jin2023text}, with less emphasis on other query types~\cite{johnson2015image}. To address the issue of limited query style adaptability in current retrieval models, we propose a novel setting: Style-diversified Query-based Image Retrieval in Fig.~\ref{fig:motivation}(b). The objective of this setting is to enable retrieval models to simultaneously accommodate various query styles, aiming to bridge the user intent gap caused by the lack of query adaptation versatility.

\begin{figure*}
    \centering
    \includegraphics[width=1.0\textwidth]{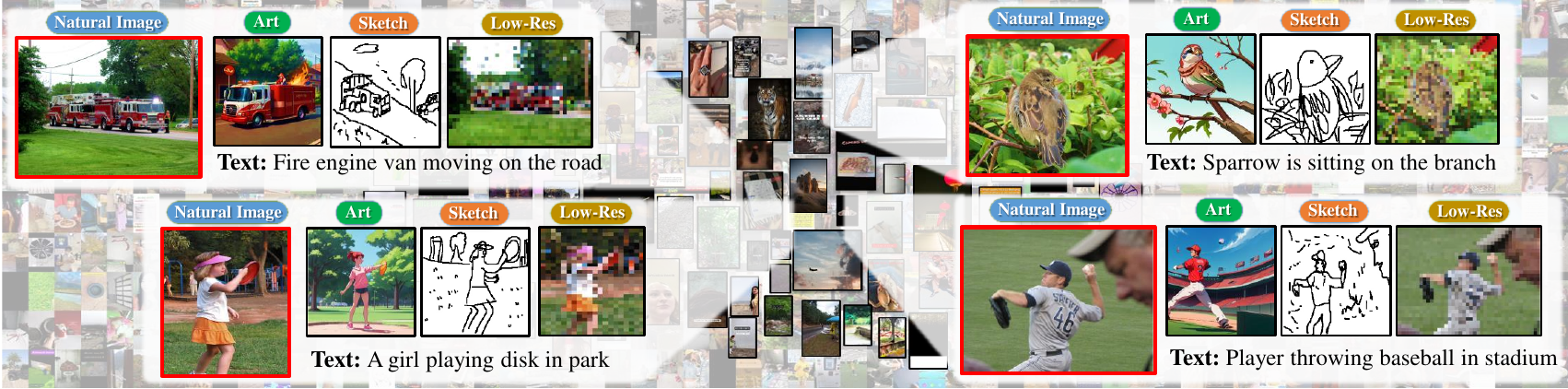}
    \vspace{-1em}
    \caption{\textbf{The Diverse-Style Retrieval Dataset~(DSR).} We propose the Diverse-Style Retrieval dataset, containing 10,000 natural images and their corresponding queries with various styles, including Sketch, Art, Low-Resolution~(Low-Res), and Text. The Diverse-Style Retrieval dataset is the first dataset for the style-diversified query-based image retrieval task.}
    \label{fig:dataset_intro}
\end{figure*}


We propose the Diverse-Style Retrieval dataset~(DSR) as the evaluation dataset of our style-diversified QBIR task. As shown in Fig.~\ref{fig:dataset_intro}, the dataset contains 10,000 natural images and four corresponding query styles: text, sketch, low-resolution, and art. \textit{(i).~Text:} the text-form query to describe the retrieval intent. \textit{(ii).~Sketch:} hand-drawn sketch by users to provide shape and pose features. \textit{(iii).~Low-Res:} users capture regions of interest from images and convert them into low-resolution images to serve as queries. \textit{(iv).~Art:} artistic-style images as retrieval queries. 

We further propose a lightweight Plug-and-Play framework, FreestyleRet, for the style-diversified retrieval task. 
For query inputs with different styles, we borrow the idea from image style transfer, calculating each query's Gram Matrix~\cite{bossett2021emotion,li2017universal} as the query's style representation, due to the Gram Matrix's ability to capture the textural information and spatial relationships between channels in the input image.
Then, we construct the high-dimensional style space by clustering all-style queries' gram matrices and taking the clustering centers as the style basis in the style space. 
With the well-constructed style space, we introduce the application of a style-init prompt tuning module on a frozen visual encoder~\cite{radford2021learning,li2022blip}, thereby enabling the encoder to adapt to various-style queries in a cost-effective manner. Specifically, given a query input, we employ its corresponding Gram matrix in conjunction with the weighted projections within the style space onto the diverse style basis as the initialization mechanism for prompt tokens in the prompt tuning procedure. 
Finally, we use the query feature from the visual encoder for further retrieval.


The proposed framework has three compelling advantages: 
\textbf{First,} The style-space construction and the style-init prompt tuning strategy enable the framework to adapt to various query styles. Experimental results on two benchmark datasets demonstrate the advantages of our model in Fig.~\ref{fig:motivation}(c).
\textbf{Second,} Our framework is compatible with the retrieval of multiple query types simultaneously, thereby promoting the single-query retrieval performance.
\textbf{Third,} the prompt-tuning structure lowers the computation cost and achieves plug-and-play abilities on various pre-trained visual encoders. 
The main contributions are as follows:
\begin{itemize}
\setlength{\itemsep}{0pt} 
    \item We are the first to propose the style-diversified QBIR task and its corresponding dataset, DSR, to address the users' intent gap problem in retrieval applications.

    \item Our framework is lightweight and plug-and-play. With the style space construction module and the style-init prompt tuning module, our framework achieves excellent performance when retrieving style-diversified queries.

    \item More encouragingly, the style-diversified queries can be simultaneously retrieved in our framework and mutually enhance each other's performance, which may have a far-reaching impact on the retrieval community.
\end{itemize}

%% file: sec/2_related.tex
\section{Related Works}
\label{sec:related_work}


\noindent\textbf{Query-based Image Retrieval. } 
Query-based Image Retrieval~(QBIR)~\cite{thomee2012interactive} aims to retrieve relevant images from a large database based on a given query. In QBIR, the query can take different forms. The earliest query form is images including natural-image retrieval~\cite{datta2008image} and face retrieval~\cite{kafai2014discrete}. With the development of cross-modal representation learning, text-style query tasks are extensively investigated, including text-image retrieval~\cite{radford2021learning,li2022blip} and text-video retrieval~\cite{jin2023diffusionret,jin2023text}. Limited research incorporates other query styles such as sketch~\cite{chowdhury2022fs,chowdhury2023scenetrilogy} and scene graph~\cite{johnson2015image}. In Fig.~\ref{fig:motivation}(b), to address the issue of single query modality being insufficient to express user search intent, we propose the style-diversified query-based image retrieval task, retrieving text, sketch, art, and low-resolution queries simultaneously.

\noindent\textbf{Promp Tuning. }
The objective of Prompt Tuning~\cite{li2021prefix,lester2021power} is to enhance the transferability of pre-trained models to downstream tasks in a cost-effective manner by incorporating learnable tokens into the fixed pre-trained models. Prompt Tuning was first proposed as text-prompt~\cite{brown2020language,liu2021p} in the language model and gained popularity in 2D~\cite{zhou2022conditional} and 3D~\cite{zha2023instance} image models. Specifically, CLIP~\cite{radford2021learning} applies fixed class-specific text as prompts. Then, CoOP~\cite{zhou2022learning} learns class-specific continuous prompts. VPT~\cite{jia2022vpt} first applies continuous prompt tokens to vision pre-trained models. Inspired by VPT, we establish a style-init prompt tuning framework for the style-diversified QBIR task.

\noindent\textbf{Image Style Transfer. }
Image style transfer~\cite{zhao2020survey,LIU2019465,cai2023image} involves the transformation of an input image to adopt the artistic style or visual characteristics of a reference image while preserving its content. Early approaches~\cite{efros1999texture,efros2023image,lee2010directional} in image style transfer are optimization-based methods relying on handcrafted features focusing only on low-level image features.
With the advent of deep learning, CNN and GAN models~\cite{richardson2021encoding,karras2019style} can extract high-level semantic features to facilitate the high-level image synthesis~\cite{gatys2016image}. For style transfer models, the Gram Matrix plays a crucial role in providing the representation of the textural and style information present in an image~\cite{bossett2021emotion,li2017universal}. We borrow the Gram Matrix for our style-diversified QBIR task, applying the Gram Matrix feature as the prompt token initialization when prompt tuning the visual encoder. 


%% file: sec/3_method.tex
\section{Preliminary of Style-Diversified QBIR}
In this section, we first introduce the problem setting of the Diverse-style Query-based Image Retrieval task in Sec.{\color{red}\ref{problem_setting}}, then introduce the new dataset we propose for this new retrieval setting in Sec.{\color{red}\ref{new_dataset}}.

\subsection{Problem Setting of Style-Diversified QBIR}
\label{problem_setting}

Given a gallery of natural images $N_I$ and a query $q_i$ from the style-specific query set $Q_s$. The goal for query-based image retrieval is to rank all images $i\in N_I $ so that the image corresponding to the query $q_i$ is ranked as high as possible. For our style-diversified QBIR setting, the goal is similar, ranking all images correctly with queries for various style-specific query sets $\{Q_s\}_{s=1}^n$.


\subsection{Datasets Construction}
\label{new_dataset}


In the context of Style-diversified Query-based Image Retrieval, we adopt two datasets as evaluation metrics: the Diverse-Style Retrieval dataset~(DSR) and ImageNet-X.


\noindent\textbf{\textit{Diverse-Style Retrieval Dataset:} } A small but fine-grained dataset constructed for style-diversified QBIR. Shown in Fig.~\ref{fig:dataset_intro}, it consists of 10,000 natural images paired with corresponding queries of four styles: text, sketch, low-res, and art. \textit{(i).~Text:} the text query used to express the retrieval intent. \textit{(ii).~Sketch:} hand-drawn sketch by users to provide shape and pose features. \textit{(iii).~Low-Res:} users capture regions of interest from images and convert them into low-resolution images to serve as queries. \textit{(iv).~Art:} artistic-style images as queries. With the rise of the AIGC~\cite{chengnull,wang2022zero,yu2023freedom}, generating images of different styles has become more convenient. Therefore, based on ten thousand natural images from FSCOCO~\cite{chowdhury2022fs}, we utilize AnimateDiff~\cite{guo2023animatediff} to generate corresponding artistic style images. We employ downsampling algorithms to generate low-resolution images. As FSCOCO provides sketch images, we use Pidinet~\cite{su2021pixel} to optimize low-quality sketch images.

\noindent\textbf{\textit{ImageNet-X:} } A large but coarse-grained dataset for style-diversified QBIR. Based on ImageNet~\cite{deng2009imagenet}, ImageNet-X contains 1M natural images and their corresponding sketch-form and art-form versions. Compared to DSR, the images in ImageNet-X are simple, containing only one object. We generate the low-resolution form for images and reconstruct ImageNet-X as the dataset for style-diversified QBIR.

\begin{figure*}[tbp]
\centering
\includegraphics[width=1.0\linewidth]{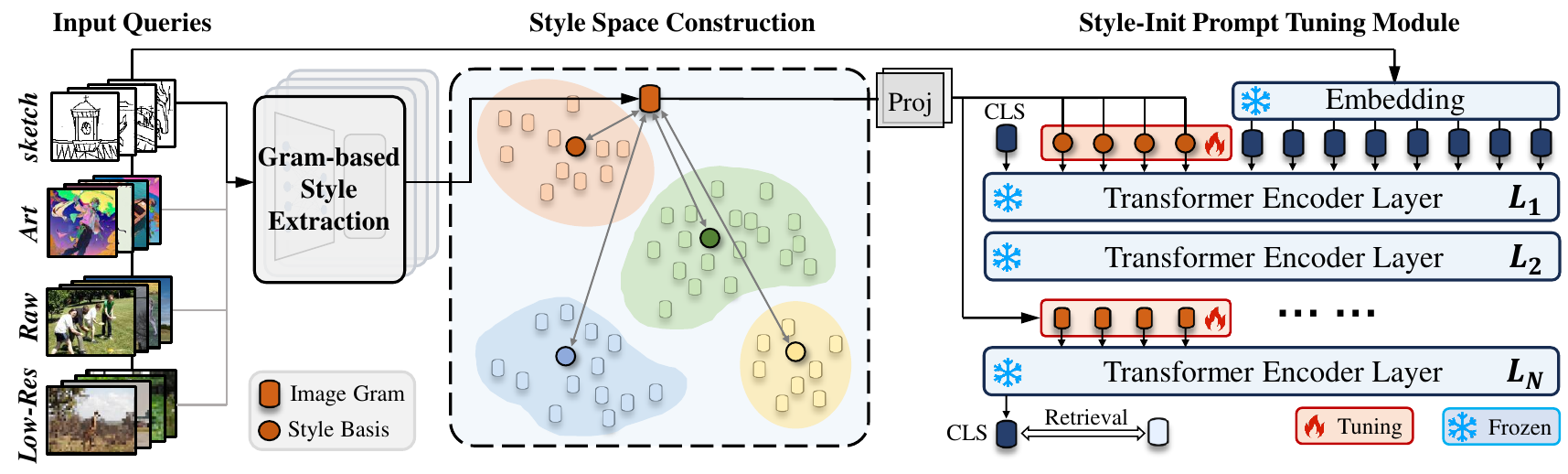}
\caption{
\textbf{The Overall Framework of our FreestyleRet.} 
For a style-diversified query input, we first extract the query's textural feature by calculating the query's gram matrix from the Gram-based Style Extraction Module. 
Then we construct the style space of queries by clustering all gram matrices and taking each clustering center as the style basis in style space. 
We further extract the query's style feature by weighted summarizing style bases based on the distance between the input query and every style basis in the style space. Finally, in the Style-Init Prompt Tuning Module, we use the gram matrix and the style feature to initialize prompt tokens, leading both textural and style information to the feature encoder for further style-diversified retrieval prediction.
}
\label{fig:framework}
\end{figure*}

\section{Methodology}
\label{sec:method}

Our model consists of three main submodules: 
(1) a \textbf{Gram-based Style Extraction Module} for generating the gram matrix of an input query, representing the query's textural feature~(Sec.{\color{red}\ref{subsec:gram-style-extraction}}).
(2) a \textbf{Style Space Construction Module} for building up the query style space by clustering queries' gram matrices and taking the cluster centers as the style basis~(Sec.{\color{red}\ref{subsec:style-space-construction}}).
(3) a \textbf{Style-Init Prompt Tuning Module} for style-specific prompt tuning a pre-trained visual encoder by initializing the prompt tokens based on the gram matrices and the style prototypes~(Sec.{\color{red}\ref{subsec:style-init-prompt-tuning}}).
The overview framework of our FreestyleRet is illustrated in Figure~\ref{fig:framework}.

\subsection{Gram-based Style Extraction Module}
\label{subsec:gram-style-extraction}


For query inputs with diverse styles, the gram-based style extraction module aims to generate the style representation from the input query. Here we borrow the style representation strategy from image style transfer, taking the gram matrix of the query's feature as the style representation.

First, we apply the frozen VGG model~\cite{simonyan2014very} to get the query's visual feature. Compared with other image feature extractors including ViT~\cite{dosovitskiy2020image} and ResNet~\cite{He7780459}, VGG is lightweight and has strong feature extraction ability during the gram matrix calculation in image style transfer works~\cite{wang2021rethinking,tao2022image}. The VGG model is constituted by a concatenation of 16 layers, consisting of stacked convolutional and fully connected layers, meticulously structured to capture complex patterns in the visual data. For query input $q_i$, we use the third convolutional layer output, shaping $112\times112\times128$, as the visual feature $v_i$ of the query $q_i$. $f_{d}(.)$ is used to downsample $v_i$.

Then, we calculate the gram matrix for query $q_i$. Specifically, the Gram Matrix $g$ of a set of vectors $t_1, ..., t_n$ in an inner product space is the Hermitian matrix of inner products: $g_{jk} = < t_j, t_k>$. $g$ represents the texture feature of vectors $t_1, ..., t_n$.
In our scenario, we calculate the gram matrix $g_i$ for $q_i$ as follows:
\begin{equation}
    g_i = (f_{d}(v_i))^{\mathsf{T}} f_{d}(v_i),
\end{equation}
where $g_i$ represents the textural feature of the query $q_i$.

\subsection{Style Space Construction Module}
\label{subsec:style-space-construction}
For style-diversified query inputs, we construct the style space $\mathbb{S}$ for queries to encode their specific styles. To generate the style-specific basis $\mathbb{B} = \{b_j\}_{j=1}^{4}$ for the style space, we cluster the gram matrices of all queries in various styles and apply each clustering center as the style-specific basis $b_j$ for the style space $\mathbb{B}$. 

During the clustering procedure, we apply the K-Means algorithm to cluster the gram matrix set $G$ for all queries from query sets in the dataset, where $G=\{g_i\}, \forall q_i\in Q_s$. We first random initialize four clustering centers $\mu_1,..., \mu4$ as the basis of the style space. Then we calculate the nearest center $c_i$ comparing each gram matrix $g_i \in G$ with existing clustering centers:
\begin{equation}
    c_i = \mathop{\arg\max}\limits_{j}||g_i - \mu_j||^2,
\end{equation}
where $j=1,...,4$. We redistribute all queries to their nearest center based on the $c_i$. Then we refine the position of $\mu_j$ by averaging all queries belong to $\mu_j$:
\begin{equation}
    \mu_j = \frac{\sum_{i=1}^m {\rm \textbf{Num}}\{c_i=j\}\times g_i}{\sum_{i=1}^m {\rm \textbf{Num}}\{c_i=j\}},
\end{equation}
We repeat the iteration of Eq.2 and Eq.3 until the clustering centers' positions converge. The well-trained clustering centers $\mu_1,..., \mu_4$ act as the style-specific basis for the constructed style space. We further use these style-specific bases to represent the style feature $s_i$ of an input query $q_i$. 
Specifically, the style feature $s_i$ is calculated by weighted summarizing all the style bases according to the cosine similarity $w$ between $q_i$ and $\mu_j, \forall j \in [1,4]$.
\begin{gather}
    w_j = \frac{e^{{\rm cos}(q_i, \mu_j)}}{\sum_{j=1}^4 e^{{\rm cos}(q_i, \mu_j)}}, \quad s_i = \sum_{j=1}^4 w_j \mu_j,
\end{gather}
The weighted summarizing calculation enables the model to generate the $q_i$'s style feature adaptively.

\newcommand{\pub}[1]{\color{gray}{\tiny{#1}}}
\newcommand{\Frst}[1]{{\textbf{#1}}}
\newcommand{\Scnd}[1]{{\underline{#1}}}
\begin{table*}[htb]
\centering
\footnotesize
\renewcommand{\arraystretch}{1.35}  
\setlength{\tabcolsep}{4.0mm}        
{
{
\begin{tabular}{l|p{80pt}|cc|cc|cc|cc}
    \toprule[1.5pt]
    \multirow{2}{*}{\textbf{\#}} & \multirow{2}{*}{\textbf{Method}} & \multicolumn{2}{c|}{\textbf{Text} \textbf{$\rightarrow$} \textbf{Image}} & \multicolumn{2}{c|}{\textbf{Sketch} \textbf{$\rightarrow$} \textbf{Image}} & \multicolumn{2}{c|}{\textbf{Art} \textbf{$\rightarrow$} \textbf{Image}} & \multicolumn{2}{c}{\textbf{Low-Res} \textbf{$\rightarrow$} \textbf{Image}} \\ 
    
    \cmidrule(rl){3-4}\cmidrule(rl){5-6}\cmidrule(rl){7-8}\cmidrule(rl){9-10}
    & & {R@1$\uparrow$} & {R@5$\uparrow$} & {R@1$\uparrow$} & {R@5$\uparrow$} & {R@1$\uparrow$} & {R@5$\uparrow$} & {R@1$\uparrow$} & {R@5$\uparrow$} \\

    \noalign{\hrule height 1.5pt}
    \rowcolor{gray!20}\multicolumn{10}{c}{\it{\textbf{Diverse-Style Retrieval Dataset}}} \\
    \hline
    1& CLIP ~\pub{ICML2021}~\cite{radford2021learning} & 66.1 & 94.7 & 47.5 & 77.3 & 58.5 & 93.7 & 45.0 & 75.7\\
    2& CLIP$^{*}$~\pub{ICML2021}~\cite{radford2021learning} & 72.2 & 96.4 & 63.6 & 93.6 & 58.2 & 90.4 & 78.8 & 97.1\\
    3& BLIP$^{*}$~\pub{ICML2022}~\cite{li2022blip} & 74.3 & 95.3 & 67.1 & 90.9 & 51.1 & 85.3 & 77.2 & 95.8\\
    4& VPT$^{*}$ ~\pub{ECCV2022}~\cite{jia2022vpt} & 69.9 & 96.1 & 73.3 & 97.0 & 66.7 & 96.5 & 81.4 & 96.0\\
    5& ImageBind~\pub{CVPR2023}~\cite{girdhar2023imagebind} & 71.0 & 95.5 & 50.8 & 79.4 & 58.2 & 86.3 & 79.0 & 96.7\\
    6& LanguageBind~\pub{Arxiv2023}~\cite{Zhu2023LanguageBindEV} & 79.7 & 98.1 & 63.6 & 89.1 & 67.5 & 92.9 & 78.6 & 94.5\\
    \hline
    \rowcolor{aliceblue!60} 7& \textbf{FreestyleRet-CLIP}  & 69.9 & 97.0 & 80.6 & \textbf{97.4} & 71.4 & \textbf{97.8} & 86.4 & 97.9\\ 
    \rowcolor{aliceblue!60} 8& \textbf{FreestyleRet-BLIP}  & \textbf{81.6} & \textbf{99.2} & \textbf{81.2} & 97.1 & \textbf{74.5} & 97.4 & \textbf{90.5} & \textbf{98.5}\\ 
    
    \noalign{\hrule height 1.5pt}
    \rowcolor{gray!20}\multicolumn{10}{c}{\it{\textbf{ImageNet-X Dataset}} \cite{deng2009imagenet}} \\
    \hline
    9& CLIP$^{*}$~\pub{ICML2021}~\cite{radford2021learning} & 42.6 & 72.7 & 41.3 & 73.9 & 38.5 & 65.3 & 74.1 & 95.7\\
    10& BLIP$^{*}$~\pub{ICML2022}~\cite{li2022blip} & 63.9 & 90.7 & 53.6 & 88.1 & 49.6 & 84.8 & 89.5 & 97.8\\
    11& VPT$^{*}$ ~\pub{ECCV2022}~\cite{jia2022vpt} & 43.3 & 85.3 & 48.6 & 84.2 & 41.6 & 88.5 & 72.7 & 89.3\\
    12& ImageBind~\pub{CVPR2023}~\cite{girdhar2023imagebind} & 57.3 & 89.7 & 53.6 & 86.2 & 49.8 & 79.3 & 81.2 & 94.3\\
    13& LanguageBind~\pub{Arxiv2023}~\cite{Zhu2023LanguageBindEV} & 68.9 & 92.3 & 62.0 & 91.5 & 60.3 & 89.9 & 87.4 & 99.5\\
    \hline
    \rowcolor{aliceblue!60} 14& \textbf{FreestyleRet-CLIP}  & 62.6 & 94.3 & 57.4 & 88.2 & 56.4 & 90.2 & 69.4 & 94.5\\
    \rowcolor{aliceblue!60} 15& \textbf{FreestyleRet-BLIP}  & \textbf{74.9} & \textbf{96.3} & \textbf{74.6} & \textbf{93.3} & \textbf{71.2} & \textbf{96.5} & \textbf{97.5} & \textbf{99.7}\\ 
 \bottomrule[1.5pt]
\end{tabular}
}
}
\vspace{-1pt}
\caption{
\textbf{Retrieval performance for Style-Diversified QBIR task.} 
We evaluate the R@1 and R@5 metrics on two benchmark datasets, the Diverse-Style Retrieval dataset and the ImageNet-X dataset. 
``$\uparrow$'' denotes that higher is better. The two forms of our FreestyleRet framework, FreestyleRet-CLIP and FreestyleRet-BLIP, outperform in multiple scenarios with different query styles compared with other baselines including cross-modality models~(CLIP, BLIP, VPT) and multimodality models~(ImageBind, LanguageBind).}
\label{tab:main_results}
\vspace{-1pt}
\end{table*}

\subsection{Style-Init Prompt Tuning Module}
\label{subsec:style-init-prompt-tuning}

To build up a lightweight and plug-and-play framework, we apply the prompt tuning procedure on a frozen pre-trained visual encoder to make the frozen visual encoder understand the various-style query inputs. As shown in Fig.~\ref{fig:framework}, during the prompt tuning, we insert four trainable prompt tokens into both the shallow layer and the bottom layer of the vision transformer encoder, to tune the visual encoder comprehensively. The prompt tokens are introduced to every transformer layer's input space. For $i$-th Layer $L_i$ in the transformer, we denote the collection of input learnable prompts $P_i$ as 
\begin{equation}
    P_i = \{p_i^k \in \mathbb{R}^d | k\in \mathbb{N}, 1\leq k \leq m\},
\end{equation}
where $d=1024$ represents the token dimension in the transformer layer. $m=4$ represents the prompt token number for each transformer layer. The style-init prompt tuning module for ViT is formulated as follows:
\begin{gather}
    [x_i, \_, E_i] = L_i(x_{i-1}, P_{i-1}, E_{i-1}), i=1,...,n \\
    f_i = \textbf{Head}(x_n),
\end{gather}
where $n$ represents the transformer layer number, $x_i$ represents the $\rm{[CLS]}$'s embedding at $L_{i}$'s input space, $E_i$ is $q_i$'s image patch embeddings. $\textbf{Head}$ represents the MLP to generate visual feature $f_i$ using the $\rm{[CLS]}$ embedding of $q_i$.

To further lead the style information to the visual encoder, given an input query $q_i$, we initialize the prompt tokens in the shallow layer based on the gram matrix from Eq.1 and initialize the tokens in the deep layer based on the style feature $s_i$ calculated from Eq.4. Further experimental analysis in Table.~\ref{ablation:prompt} shows that differentiated style initialization across different layers can boost the performance of the ViT-based visual encoder.

\subsection{Training and Inference}
\label{subsec:train-infer}

As shown in Fig.~\ref{fig:framework}, our FreestyleRet iterates the dataset twice during the training process. We first construct the style space during the first iteration. Then we apply the well-constructed style space for style-init prompt tuning during the second iteration. The overall loss $\mathcal{L}$ of our model is the triplet loss:
\begin{gather}
    {\rm \textbf{dist}}(x, y) = 1 - {\rm cos}(x, y), \\
    \mathcal{L} = \frac{1}{B}\sum_{i=1}^B({\rm max}(0, {\rm \textbf{dist}}(F_i, P_i) - {\rm \textbf{dist}}(F_i, N_i)+\alpha))
\end{gather}
where $F$ represent the image features $F=\{f_i\}_1^n$. $P$ represents the positive samples and $N$ represents the negative samples. During the training, we take the ground-truth retrieval answer as $P$. For $N$ we randomly select another image from the same query-style set as $q_i$. We set the hyperparameter $\alpha$ to 1.0.

Our inference process iterates the test dataset once, using the gram-based style extraction module and the well-constructed style space to get the textural feature from the gram matrix and style feature for the input query. Then we apply the style-init prompt tuning module for retrieval.

%% file: sec/4_exp.tex
\section{Experiments} 

\subsection{Experimental Settings}
For the experiments on the DSR and the ImageNet-X datasets, FreestyleRet is trained on one A100 GPU with batch size 24 and 20 training epochs. The learning rate is set to 1e-5 and is linearly warmed up in the first epochs and then decayed by the cosine learning rate schedule. 
During training, all input images are resized into $224\times224$ resolution and then augmented by normalized operation.

\begin{table*}[htb]
\centering
\footnotesize
\renewcommand{\arraystretch}{1.35}  
\setlength{\tabcolsep}{1.7mm}        
{
{
\begin{tabular}{l|p{70pt}|cc|cc|cc}
    \toprule[1.5pt]
    \multirow{2}{*}{\textbf{\#}} & \multirow{2}{*}{\textbf{Method}} & \multicolumn{2}{c|}{\textbf{Sketch}} & \multicolumn{2}{c|}{\textbf{Art}} & \multicolumn{2}{c}{\textbf{Low-Resolution}}\\ 
    
    \cmidrule(rl){3-4}\cmidrule(rl){5-6}\cmidrule(rl){7-8}
    & & Text\textbf{$\rightarrow$}Image & Sketch+Text\textbf{$\rightarrow$}Image & Text\textbf{$\rightarrow$}Image & Art+Text\textbf{$\rightarrow$}Image & Text\textbf{$\rightarrow$}Image & Low-Res+Text\textbf{$\rightarrow$}Image  \\

    \noalign{\hrule height 1.5pt}
    1& CLIP$^{*}$ & 72.2 & 65.0{$_{\textcolor{black}{(-7.2)}}$} & 72.2 & 57.8{$_{\textcolor{black}{(-14.4)}}$} & 72.2 & 84.7{$_{\textcolor{red}{(+12.5)}}$}\\
    2& BLIP$^{*}$ & 74.3 & 74.2{$_{\textcolor{black}{(-0.1)}}$} & 74.3 & 58.3{$_{\textcolor{black}{(-16.0)}}$} & 74.3 & \textbf{88.3}{$_{\textcolor{red}{(+14.0)}}$}\\
    \hline
    \rowcolor{aliceblue!60} 3& \textbf{FreestyleRet} & 69.9 & \textbf{82.5}{$_{\textcolor{red}{(+12.6)}}$} & 69.9 & \textbf{76.6}{$_{\textcolor{red}{(+6.7)}}$} & 69.9 & \textbf{86.7}{$_{\textcolor{red}{(+16.8)}}$}\\
 \bottomrule[1.5pt]
\end{tabular}
}
}
\vspace{-1pt}
\caption{
\textbf{Retrieval performance with multi-style queries simultaneously.} 
The additional query inputs~(sketch, art, low-res) can boost the text-image retrieval capability in our FreestyleRet while showing a negative influence on baseline models, including CLIP and BLIP.
}
\label{tab:multi-query}
\vspace{-1pt}
\end{table*}

\begin{figure*}
    \begin{minipage}[b]{.65\linewidth}
        \centering
        \footnotesize
        \renewcommand{\arraystretch}{1.35}  
        \setlength{\tabcolsep}{2.5mm}        
        \begin{tabular}{l|ccc|c|c|c}
            \toprule[1.5pt]
            \textbf{\#} & \textbf{Shallow} & \textbf{Bottom} & \textbf{Token-Num} & \textbf{Sketch\textbf{$\rightarrow$}I} & \textbf{Art\textbf{$\rightarrow$}I} & \textbf{Low-Res\textbf{$\rightarrow$}I} \\ 
            \noalign{\hrule height 1.5pt}
            1 & Random & Random & 4 & 68.1 & 63.5 & 78.8\\
            2 & Style Space & Random & 4 & 76.7 & 69.1 & 82.4\\
            3 & Random & Gram & 4 & 76.8 & 69.2 & 81.8\\
            4 & Gram & Style Space & 4 & 78.1 & 69.5 & 84.4\\
            \rowcolor{aliceblue!60} 5 & Style Space & Gram & 4 & \textbf{80.6} & \textbf{71.4} & \textbf{86.4}\\
            \midrule
            6 & Style Space & Gram & 1 & 68.2 & 64.7 & 79.1\\
            7 & Style Space & Gram & 2 & 72.3 & 65.9 & 82.8\\
            8 & Style Space & Gram & 8 & 77.9 & 67.1 & 80.7\\
            \bottomrule[1.5pt]
        \end{tabular}
        \captionof{table}{
        \textbf{The analysis for the prompt token design.} We ablate the prompt tokens' number, insert position, and initialization feature in our FreestyleRet framework.
        }
        \label{ablation:prompt}
    \end{minipage}
    \quad \quad
    \begin{minipage}[b]{.3\linewidth}
        \centering
        \includegraphics[width=0.88\linewidth]{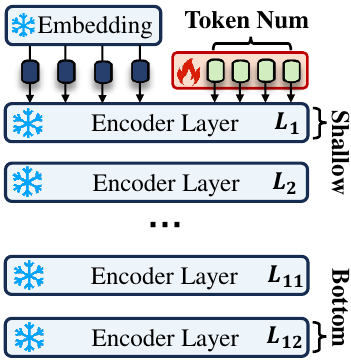}
        \caption{The prompt tuning structure in the Freestyle framework.}
        \label{fig:prompt_structure}
    \end{minipage}
\end{figure*}

\subsection{Main Results}
we select the most recent multi-modality pre-trained models for comparison, including three cross-modality pre-trained models~(CLIP, BLIP, VPT) and two multi-modality pre-trained models~(ImageBind, LanguageBind). 
Specifically, we prompt-tuning the cross-modality models to adapt style-diversified inputs. $\textbf{*}$ represents the prompt-tuning version of the vanilla models. As for the multi-modality pre-trained models, we evaluate the zero-shot performance on the sty-diversified retrieval task due to multi-modality models' comprehensionability on multi-style image inputs. We apply two benchmark datasets, including the ImageNet-X and the DSR dataset, for our style-diversified retrieval task. The results in Table.~\ref{tab:main_results} yield three observations:

\begin{figure*}[tbp]
\centering
\includegraphics[width=1.0\linewidth]{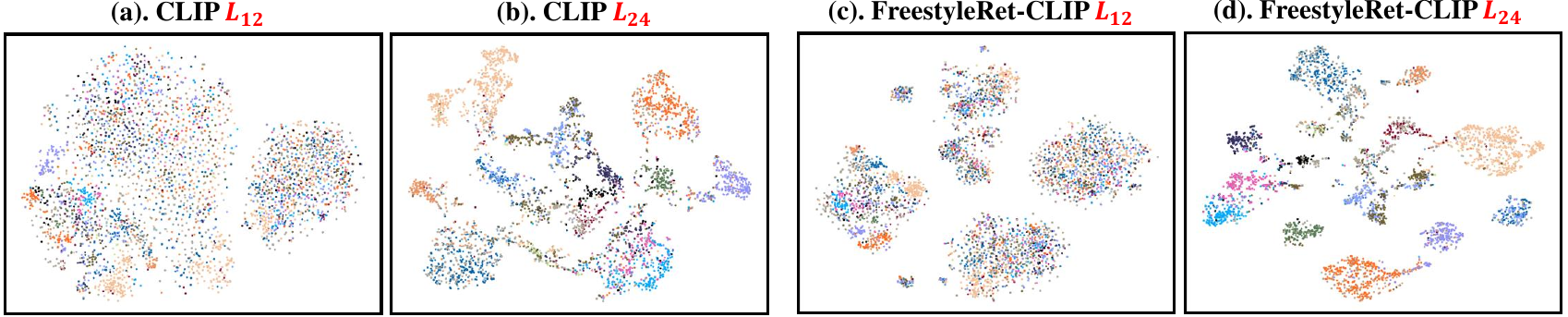}
\caption{
\textbf{The Feature Distribution Analysis for our FreestyleRet.} We make t-SNE~\cite{van2009learning} visualization for the middle layer~$L_{12}$ and the output layer~$L_{24}$ from the FreestyleRet and the CLIP baseline. The colorful dots donate query features, different colors represent different semantics. Compared to the baseline, FreestyleRet achieves better semantic clustering at both middle and deep layers.
}
\label{fig:tsne}
\vspace{-5pt}
\end{figure*}


\textbf{\textit{(i). Cross-modality and Multi-modality models have the potential for improvement in the style-diversified retrieval task.}} 
Line.1 in Tab.~\ref{tab:main_results} shows that zero-shot CLIP performs badly compared with our FreestyleRet. This limitation arises from the inability of vision-linguistic models like CLIP to distinguish visual inputs with different styles from those of natural images in the feature space. With the prompt tuning process, cross-modality models have significant improvements, as shown in line.2-4 and line.9-11. As for the multi-modality models, ImageBind and LanguageBind, line.5-6 and line.12-13 show that multi-modality models have style-diversified retrieval abilities.

\textbf{\textit{(ii). The CLIP-form and Blip-form models of our FreestyleRet framework outperform both cross-modality and multi-modality models.}}
Claimed in Sec.\ref{subsec:style-init-prompt-tuning}, our FreestyleRet is a plug-and-play framework that can easily applied to various pretrained visual encoders. Here we apply our FreestyleRet on two ViT-based visual encoders from CLIP and BLIP. We use FreestyleRet-CLIP and FreestyleRet-BLIP as the generated models. Line.7-8 and line.14-15 in Tab.~\ref{tab:main_results} show that both FreestyleRet-CLIP and FreestyleRet-BLIP outperform the cross-modality and multi-modality baselines, demonstrating the effectiveness of our plug-and-play framework.

\textbf{\textit{(iii). In our FreestyleRet framework,  style-diversified queries can be simultaneously retrieved and mutually enhance the text-image retrieval performance.}}
As shown in Tab.~\ref{tab:multi-query}, when conducting text-image retrieval, the additional query inputs~(sketch, art, low-res) can significantly boost the text-image retrieval capability of our FreestyleRet framework. However, for baseline models, the additional query signals cannot stably improve the text-image retrieval performance. In line.1-2 in Tab.~\ref{tab:multi-query} the additional sketch and art queries have a negative effect on the CLIP and BLIP.

\begin{table}[]
    \centering
    \footnotesize
    \renewcommand{\arraystretch}{1.3}  
    \setlength{\tabcolsep}{3.0mm}        
    \begin{tabular}{l|cc}
        \toprule[1.5pt]
        \textbf{Method} & \textbf{Parameters(M)} & \textbf{Speed(ms)}\\
        \noalign{\hrule height 1.5pt}
        CLIP~\pub{ICML2021}~\cite{radford2021learning} & 427M & 68ms\\
        BLIP~\pub{ICML2022}~\cite{li2022blip} & 891M & 62ms\\
        VPT~\pub{ECCV2022}~\cite{jia2022vpt} & 428M & 73ms\\
        ImageBind~\pub{CVPR2023}~\cite{jia2022vpt} & 1200M & 372ms\\
        \hline
        \rowcolor{aliceblue!60} \textbf{FreestyleRet-CLIP} & 476M{$_{\textcolor{red}{(+29)}}$} & 96ms{$_{\textcolor{red}{(+28)}}$}\\ 
        \rowcolor{aliceblue!60} \textbf{FreestyleRet-BLIP} & 940M{$_{\textcolor{red}{(+29)}}$} & 101ms{$_{\textcolor{red}{(+39)}}$}\\ 
        \bottomrule[1.5pt]
    \end{tabular}
    \caption{Comparison of the computation complexity between our FreestyleRet and baselines. Our framework is computationally efficient from the trainable parameter and inference speed aspects.}
    \label{tab:computation}
\end{table}
\vspace{-3pt}

\subsection{Ablation Studies}
In this section, we ablate the detailed performance analysis and the model design choices of our FreestyleRet framework. The details are as follows.

\vspace{-3pt}
\subsubsection{Ablation for Prompt Tuning Structure}
We ablate the prompt tuning structure in our FreestyleRet framework from three aspects: the prompt token initialization feature, the position where the prompt tokens are inserted, and the prompt token number. Table.~\ref{ablation:prompt} shows the ablation results. Furthermore, Fig.~\ref{fig:prompt_structure} proposes the detailed structure of the prompt tuning module in FreestyleRet. 

\textbf{The prompt token position. }
Previous prompt tuning models~\cite{jia2022vpt,liu2021p,liu2022p} analyzed that inserting the learnable prompt tokens in all layers in the transformer~(Deep Prompt) has better performance than in the first layer in the transformer~(Shallow Prompt). In the prompt tuning module of our FreestyleRet, we also adopt the deep prompt idea and insert all the learnable prompt tokens into all layers.

\textbf{The prompt token initialization. }
We analyze the impact of the prompt token initialization by applying different initialization strategies in different positions of the visual encoder. Line.1-5 in Table.~\ref{ablation:prompt} show the ablation results, where ``Random'' represents random initialization, ``Gram'' represents initializing with textual information from the gram matrix, and ``Style Space'' represents initializing with style information from the style space feature. The random initialization in line.1 performs worst, demonstrating that applying textural and style representation as initialization is necessary. We make various initialization attempts in line.2-4 and find that initializing the shallow-layer prompt tokens with style features, while initializing the deep-layer prompt tokens with gram matrices, achieves the best performance.

\textbf{The prompt token number. }
We make ablation studies for the number of prompt tokens that are inserted into the visual encoder during the prompt tuning stage. As shown in line.5-8 in Table.~\ref{ablation:prompt}, our FreestyleRet framework, adopting 4 prompt tokens, outperforms other number settings including 1, 2, 8 prompt tokens under three evaluation metrics.

\vspace{-1pt}
\subsubsection{Computation Comparison}
To validate the lightweight nature of our FreestyleRet framework and its ease of integration into existing retrieval models, we analyze the computational complexity of our framework compared with other baselines. Table.~\ref{tab:computation} shows the statistical analysis of trainable parameters and inference time per batch for our FreestyleRet framework and other baselines. Compared with the multi-modality model, ImageBind, our FreestyleRet is lightweight both in the trainable parameter and the inference speed. Compared with the cross-modality models, including CLIP and BLIP, our framework slightly increases the inference time and the trainable parameter while maintaining rapid deployment and application without significant impact.

\begin{figure*}[tbp]
\centering
\includegraphics[width=1.0\linewidth]{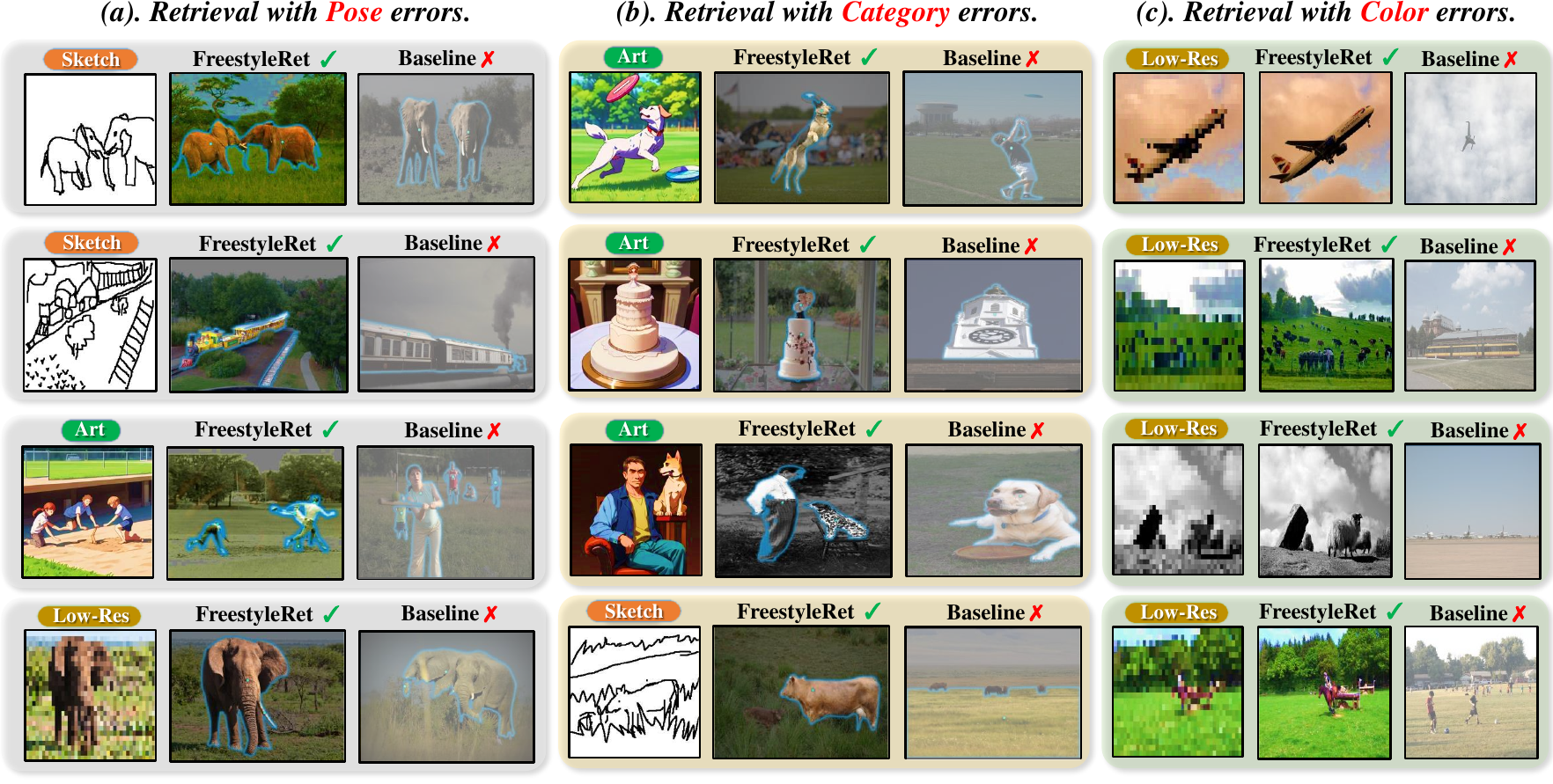}
\caption{
\textbf{The Case Study for our FreestyleRet and the CLIP baseline. } We visualize style-diversified queries and their corresponding retrieval answers from our FreestyleRet model and the baseline model. We summarize three common retrieval errors: pose errors, category errors, and color errors. Our FreestyleRet is capable of effectively retrieving based on specific pose, category, and color information from sketch, art, and low-resolution queries. 
However, the CLIP baseline model tends to encounter errors in these cases.
}
\label{fig:visualization}
\end{figure*}

\subsection{Qualitative Analysis}
In this section, we do the qualitative analysis of our framework's performance by visualizing the high-dimensional feature distribution and the prediction cases from our FreestyleRet framework compared with the baseline, the prompt tuning form of the CLIP model.  

\subsubsection{Feature Distribution Analysis}
In Fig.\ref{fig:tsne}, we analyze the feature distribution using t-SNE~\cite{van2009learning} visualizations for the middle layer~$L_{12}$ features and the output layer~$L_{24}$ features from our FreestyleRet and the prompt-tuning CLIP as the baseline. The colorful dots donate style-diversified query features, different colors represent different semantic classes, including dog, cat, truck, etc. 
Comparing Fig.~\ref{fig:tsne}~(d) with Fig.~\ref{fig:tsne}~(b), our FreestyleRet can successfully cluster together different style queries with similar semantic classes, while the baseline cannot achieve semantic clustering. Comparing Fig.~\ref{fig:tsne}~(c) with Fig.~\ref{fig:tsne}~(a), in the middle layer, our FreestyleRet has already clustered similar semantic queries, while the baseline cannot understand the semantic feature from style-diversified queries, showing random cluster in high-dimensional space.

\subsubsection{Case Study and Error Analysis}
In Fig.~\ref{fig:visualization}, we visualize the style-diversified query inputs and their corresponding retrieval answers from our FreestyleRet model and the CLIP baseline model. We summarize three common retrieval errors in the case analysis, where pose errors, category errors, and color errors represent the false retrieval result with false poses, categories, and colors. We propose the pose error cases in Fig.~\ref{fig:visualization}(a). The pose information is contained widely in different style queries. Thus, pose error cases occur in sketch, art, low-res queries. The art queries tend to reshape the category into the art form. Thus, in Fig.~\ref{fig:visualization}(b), most of the category errors occur in the art-style retrieval task. For the low-resolution query retrieval task, color is vital retrieval information. In Fig.~\ref{fig:visualization}(c), we show the color errors from the low-resolution retrieval task. 
Compared with the CLIP baseline model, our FreestyleRet framework can achieve fine-grained retrieval based on the pose, category, and color information from style-diversified query inputs, demonstrating the superiority of our FreestyleRet framework.


%% file: sec/5_conclusion.tex
\section{Conclusion}
\label{sec:conclusion}

In this paper, we are the first to propose the style-diversified query-based image retrieval task to address the issue of limited query style adaptability in current retrieval models. We construct a corresponding dataset, the Diverse-Style Retrieval dataset, for the style-diversified QBIR task. We further propose a lightweight plug-and-play framework, FreestyleRet, to retrieve from style-diversified query inputs. Our FreestyleRet extracts the query's textural and style features from the gram matrix as the style-diversified initialization for the prompt tuning stage. This facilitates the framework in adapting to style-diversified query-based image retrieval. Experiment results show the effectiveness and computational efficiency of our FreestyleRet. In future work, we will incorporate a broader range of query styles into our Diversified-Style Dataset and explore more efficient style-based prompt-tuning strategies for our framework. 

%% file: sec/6_suppl.tex
\clearpage
\setcounter{page}{1}
\maketitlesupplementary

\renewcommand*\contentsname{Supplementary Summary}

\begin{table}[!h]
    \centering
    \footnotesize
    \renewcommand{\arraystretch}{1.35}  
    \setlength{\tabcolsep}{2.5mm}        
    \begin{tabular}{l|cc|c|c|c}
        \toprule[1.5pt]
        \textbf{\#} & \textbf{Shallow} & \textbf{Bottom} & \textbf{S\textbf{$\rightarrow$}I} & \textbf{A\textbf{$\rightarrow$}I} & \textbf{LR\textbf{$\rightarrow$}I} \\ 
        \noalign{\hrule height 1.5pt}
        1 & Random & - & 78.1 & 70.2 & 85.3\\
        2 & Style Space & - & 78.5 & 71.6 & 85.5\\
        3 & Gram Matrix & - & 79.0 & 70.9 & 84.2\\
        \midrule
        4 & - & Random & 78.5 & 70.3 & 83.5\\
        5 & - & Style Space & 79.4 & 70.7 & 84.7\\
        6 & - & Gram Matrix& 79.2 & 70.8 & 83.8\\
        \midrule
        \rowcolor{aliceblue!60} 7 & Style Space & Gram & \textbf{80.6} & \textbf{71.4} & \textbf{86.4}\\
        \bottomrule[1.5pt]
    \end{tabular}
    \captionof{table}{
    \textbf{The ablation analysis for the prompt token inserting strategy.} We ablate three prompt-token inserting strategies in our FreestyleRet framework, including inserting in the shallow layer, inserting in the deep layer, and inserting in both layers. Experiments show that inserting in both shallow and deep layers achieves the best performance.
    }
    \label{ablation:prompt}
\end{table}

\section{Supplements for Experimental Results}
We present the supplementary experiments for style-diversified retrieval results and the ablation studies for our FreestyleRet framework.

\subsection{Extra Ablation for Prompt Token Inserting Strategies}
In the main paper, we conducted ablation experiments on the initialization choices and the number of prompt-tuning tokens in the prompt-tuning structure. In the supplementary material, we further performed ablation on the number of layers in the prompt tuning structure. Specifically, in Table.~\ref{ablation:prompt}, we compared the performance of the model when only inserting prompt tokens in shallow layers, only inserting prompt tokens in deep layers, and inserting prompt tokens in both shallow and deep layers. All experiments in Table.~\ref{ablation:prompt} are conducted by our FreestyleRet framework on the DSR dataset. ``S$\rightarrow$I'' represents sketch to image retrieval. ``A$\rightarrow$I'' represents art to image retrieval. ``LR$\rightarrow$I'' represents low-resolution to image retrieval.

Compare line.7 with line.1-3 and line.4-6 in Table.~\ref{ablation:prompt}, inserting prompt tokens in both shallow and deep layers outperforms other inserting strategies.
In comparison to the random initialization method~(line.1\&4), both style initialization~(line.2\&5) and gram initialization~(line.3\&6) result in higher accuracy. Additionally, the deep-layer prompt provides the encoder with a larger bias, contributing to a slight increase in performance compared to the shallow-layer prompt strategy.  

\begin{figure}
    \centering
    \includegraphics[width=1.0\linewidth]{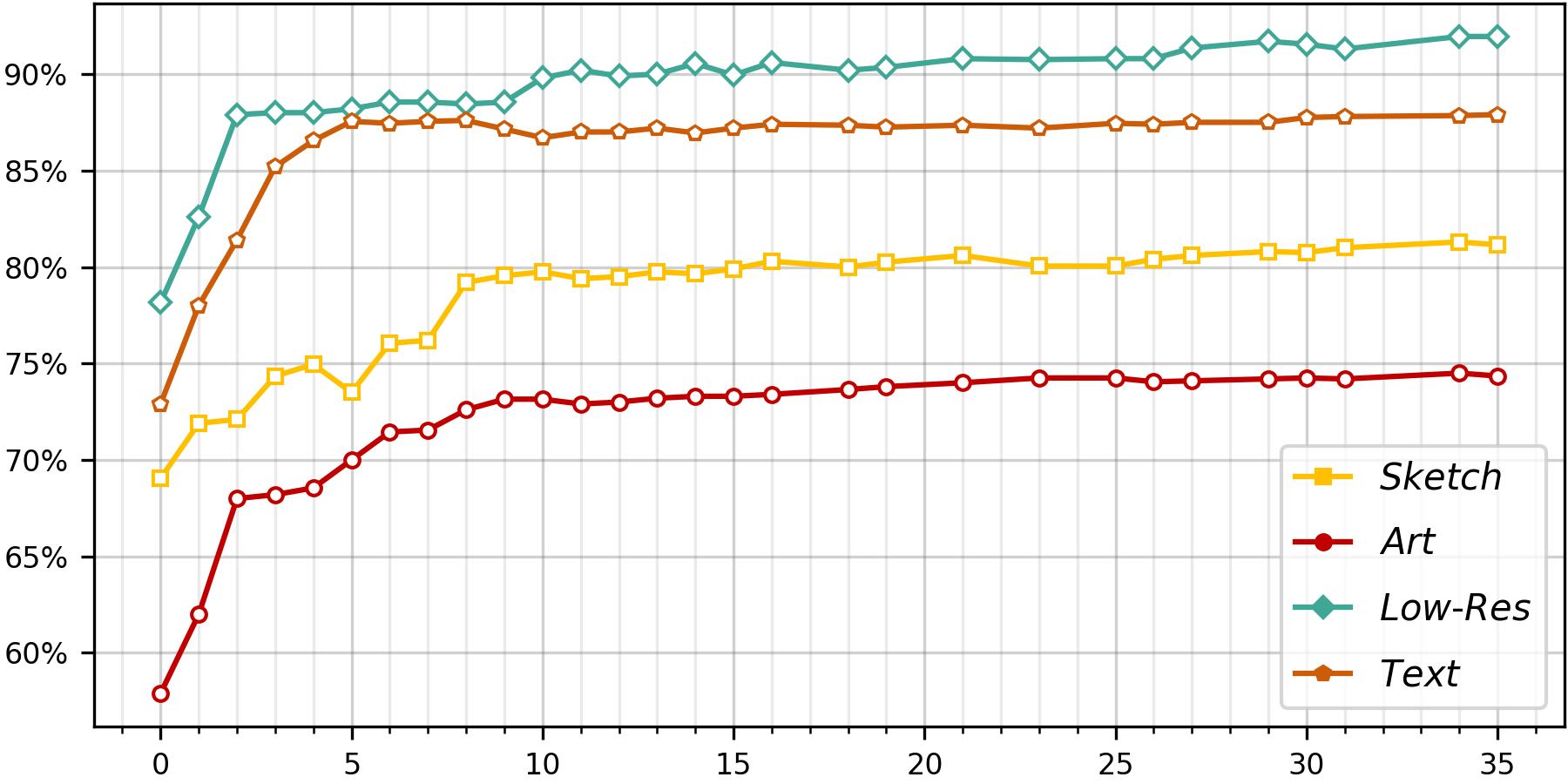}
    \caption{
    \textbf{The epoch analysis for our FreestyleRet framework.} For style-diversified retrieval tasks, our lightweight framework achieves a rather good performance under 10 epochs.
    }
    \label{fig:epoch}
\end{figure}

\subsection{Epoch Analysis for the FreestyleRet}
To demonstrate the fast convergence and low computational cost of our FreestyleRet framework, we conduct the epoch analysis for our FreestyleRet and visualize the performance change under different epochs training.

As shown in Fig.\ref{fig:epoch}, our FreestyleRet framework achieves better performance and faster convergence speed with 5-10 training epochs compared to other baselines such as prompting tuning BLIP, CLIP, and VPT models. These pre-trained baseline models need at least 50 or more training epochs to converge.

Also, we observe that text and low-resolution retrieval converge after 5 training epochs, faster than art and sketch retrieval~(10 epochs). The text modal and the low-resolution style have less information gap between the natural image modality, so their performance converges faster. On the other hand, the sketch style and the art style, containing more style and textural information, require more epochs (about 10) to achieve better retrieval accuracy. Additionally, each training epoch only takes 4 minutes. The performance in the main body is an average of epoch-5, epoch-10, and epoch-20 evaluation results.

\begin{table*}[t]
\centering
\footnotesize
\renewcommand{\arraystretch}{1.35}  
\setlength{\tabcolsep}{4.0mm}        
{
{
\begin{tabular}{l|p{80pt}|cc|cc|cc|cc}
    \toprule[1.5pt]
    \multirow{2}{*}{\textbf{\#}} & \multirow{2}{*}{\textbf{Method}} & \multicolumn{2}{c|}{Image\textbf{$\rightarrow$}Text} & \multicolumn{2}{c|}{Sketch\textbf{$\rightarrow$}Text} & \multicolumn{2}{c|}{Art\textbf{$\rightarrow$}Text} & \multicolumn{2}{c}{Low-Res\textbf{$\rightarrow$}Text} \\ 
    
    \cmidrule(rl){3-4}\cmidrule(rl){5-6}\cmidrule(rl){7-8}\cmidrule(rl){9-10}
    & & {R@1} & {R@5} & {R@1} & {R@5} & {R@1} & {R@5} & {R@1} & {R@5} \\

    \noalign{\hrule height 1.5pt}
    1& CLIP$^{*}$ & 55.2 & 90.8 & 48.4 & 87.9 & 64.5 & 96.8 & 42.6 & 81.8 \\
    2& BLIP$^{*}$ & 71.4 & 94.9 & 55.5 & 87.0 & 81.0 & 98.8 & 49.2 & 81.8\\
    3& VPT & 52.2 & 91.7 & 45.2 & 87.7 & 52.7 & 94.6 & 44.3 & 84.6\\
    4& ImageBind & 73.5 & 96.5 & 56.1 & 88.4 & 82.7 & 99.0 & 42.4 & 73.8\\
    5& LanguageBind & 80.5 & 98.3 & 63.9 & 91.6 & 87.2 & 99.7 & 56.9 & 86.8\\
    \midrule
    \rowcolor{aliceblue!60} 6& FreestyleRet-CLIP & 71.6 & 98.0 & 66.7 & \textbf{96.7} & 74.4 & 99.1 & 64.1 & 94.8\\
    \rowcolor{aliceblue!60} 7& FreestyleRet-BLIP & \textbf{82.8} & \textbf{99.0} & \textbf{71.0} & 96.4 & \textbf{86.6} & \textbf{99.7} & \textbf{69.5} & \textbf{96.9}\\
 \bottomrule[1.5pt]
\end{tabular}
}
}
\caption{
\textbf{The Text-Retrieval performance of our FreestyleRet and baseline models.} 
}
\label{tab:result_text}
\end{table*}

\begin{table*}[h]
\centering
\footnotesize
\renewcommand{\arraystretch}{1.35}  
\setlength{\tabcolsep}{4.0mm}        
{
{
\begin{tabular}{l|p{80pt}|cc|cc|cc|cc}
    \toprule[1.5pt]
    \multirow{2}{*}{\textbf{\#}} & \multirow{2}{*}{\textbf{Method}} & \multicolumn{2}{c|}{Image\textbf{$\rightarrow$}Art} & \multicolumn{2}{c|}{Sketch\textbf{$\rightarrow$}Art} & \multicolumn{2}{c|}{Text\textbf{$\rightarrow$}Art} & \multicolumn{2}{c}{Low-Res\textbf{$\rightarrow$}Art} \\ 
    
    \cmidrule(rl){3-4}\cmidrule(rl){5-6}\cmidrule(rl){7-8}\cmidrule(rl){9-10}
    & & {R@1} & {R@5} & {R@1} & {R@5} & {R@1} & {R@5} & {R@1} & {R@5} \\

    \noalign{\hrule height 1.5pt}
    1& CLIP$^{*}$ & 63.0 & 94.7 & 61.2 & 92.7 & 75.5 & 98.2 & 51.9 & 87.9 \\
    2& BLIP$^{*}$ & 57.1 & 88.5 & 44.8 & 82.8 & 82.8 & 98.7 & 39.4 & 79.3\\
    3& VPT & 67.4 & 95.5 & 60.3 & 93.1 & 61.6 & 96.5 & 44.3 & 84.6\\
    4& ImageBind & 46.4 & 80.5 & 28.7 & 60.8 & 82.6 & 98.9 & 57.8 & 89.6\\
    5& LanguageBind & 65.8 & 93.2 & 41.1 & 77.7 & 86.7 & 99.2 & 34.8 & 72.0\\
    \midrule
    \rowcolor{aliceblue!60} 6& FreestyleRet-CLIP & 72.9 & \textbf{97.8} & \textbf{66.5} & \textbf{96.2} & 85.0 & 99.6 & \textbf{62.8} & \textbf{94.1}\\
    \rowcolor{aliceblue!60} 7& FreestyleRet-BLIP & \textbf{73.6} & 97.4 & 63.1 & 94.4 & \textbf{90.2} & \textbf{99.7} & 60.1 & 92.2\\
 \bottomrule[1.5pt]
\end{tabular}
}
}
\caption{
\textbf{The Art-Retrieval performance of our FreestyleRet and baseline models.} 
}
\label{tab:result_art}
\end{table*}

\subsection{More Experimental Results for the Style-Diversified Retrieval Task}
In order to comprehensively validate the superiority of our FreestyleRet model in handling the retrieval of queries with different styles, we conducted extensive experiments involving cross-modal retrieval among various style-diversified queries, including any queries to Text modality, any queries to Art modality, any queries to Sketch modality, and any queries to Low-resolution modality. 

We present the performance comparison between our FreestyleRet and other baselines in Table.~\ref{tab:result_text}~(Any$\rightarrow$Text), Table.~\ref{tab:result_art}~(Any$\rightarrow$Art), Table.~\ref{tab:result_sketch}~(Any$\rightarrow$Sketch), and Table.~\ref{tab:result_Low_Res}~(Any$\rightarrow$Low-resolution Images). All experiments are conducted on the DSR dataset. Experimental results demonstrate that our FreestyleRet framework achieves state-of-the-art~(SOTA) performance in almost all retrieval scenarios. Specifically, in complex scenarios including sketch and art style retrieval, our FreestyleRet model outperforms other baseline models by a significant margin of 6\%-10\% due to the integration of our style extraction module and style-based prompt tuning module.

In Table.~\ref{tab:result_Low_Res}, we observed that the fine-tuned BLIP model outperforms our FreestyleRet model in the retrieval of Images to low-resolution images. This is because there is a high semantic similarity between low-resolution images and natural images, and simple prompt tuning allows the baseline model to achieve good results. However, our model still surpasses the baseline in tasks involving cross-modal retrieval from other modalities to low-resolution image modalities.

\begin{table*}[htb]
\centering
\footnotesize
\renewcommand{\arraystretch}{1.35}  
\setlength{\tabcolsep}{4.0mm}        
{
{
\begin{tabular}{l|p{80pt}|cc|cc|cc|cc}
    \toprule[1.5pt]
    \multirow{2}{*}{\textbf{\#}} & \multirow{2}{*}{\textbf{Method}} & \multicolumn{2}{c|}{Image\textbf{$\rightarrow$}Sketch} & \multicolumn{2}{c|}{Art\textbf{$\rightarrow$}Sketch} & \multicolumn{2}{c|}{Text\textbf{$\rightarrow$}Sketch} & \multicolumn{2}{c}{Low-Res\textbf{$\rightarrow$}Sketch} \\ 
    
    \cmidrule(rl){3-4}\cmidrule(rl){5-6}\cmidrule(rl){7-8}\cmidrule(rl){9-10}
    & & {R@1} & {R@5} & {R@1} & {R@5} & {R@1} & {R@5} & {R@1} & {R@5} \\

    \noalign{\hrule height 1.5pt}
    1& CLIP$^{*}$ & 70.5 & 96.1 & 60.5 & 92.9 & 55.0 & 90.8 & 60.4 & 90.9 \\
    2& BLIP$^{*}$ & 69.8 & 93.5 & 47.6 & 82.8 & 58.6 & 89.8 & 52.3 & 82.8\\
    3& VPT & 71.7 & 96.2 & 62.3 & 92.9 & 49.4 & 88.6 & 63.3 & 91.5\\
    4& ImageBind & 54.0 & 81.8 & 38.3 & 71.6 & 56.1 & 88.4 & 26.2 & 52.5\\
    5& LanguageBind & 74.6 & 96.1 & 57.5 & 87.0 & 65.7 & 94.0 & 54.5 & 83.8\\
    \midrule
    \rowcolor{aliceblue!60} 6& FreestyleRet-CLIP & 77.8 & \textbf{98.1} & 66.5 & \textbf{96.2} & 72.3 & 97.4 & 68.7 & \textbf{95.1}\\
    \rowcolor{aliceblue!60} 7& FreestyleRet-BLIP & \textbf{80.5} & 97.7 & \textbf{66.8} & 94.9 & \textbf{76.6} & \textbf{97.7} & \textbf{71.1} & 94.3\\
 \bottomrule[1.5pt]
\end{tabular}
}
}
\caption{
\textbf{The Sketch-Retrieval performance of our FreestyleRet and baseline models.} 
}
\label{tab:result_sketch}
\end{table*}

\begin{table*}[htb]
\centering
\footnotesize
\renewcommand{\arraystretch}{1.35}  
\setlength{\tabcolsep}{4.0mm}        
{
{
\begin{tabular}{l|p{80pt}|cc|cc|cc|cc}
    \toprule[1.5pt]
    \multirow{2}{*}{\textbf{\#}} & \multirow{2}{*}{\textbf{Method}} & \multicolumn{2}{c|}{Image\textbf{$\rightarrow$}Low-Res} & \multicolumn{2}{c|}{Art\textbf{$\rightarrow$}Low-Res} & \multicolumn{2}{c|}{Text\textbf{$\rightarrow$}Low-Res} & \multicolumn{2}{c}{Sketch\textbf{$\rightarrow$}Low-Res} \\ 
    
    \cmidrule(rl){3-4}\cmidrule(rl){5-6}\cmidrule(rl){7-8}\cmidrule(rl){9-10}
    & & {R@1} & {R@5} & {R@1} & {R@5} & {R@1} & {R@5} & {R@1} & {R@5} \\

    \noalign{\hrule height 1.5pt}
    1& CLIP$^{*}$ & 79.3 & 97.2 & 53.0 & 89.2 & 46.0 & 82.3 & 59.5 & 92.4 \\
    2& BLIP$^{*}$ & \textbf{89.0} & 40.8 & 73.9 & 87.0 & 51.5 & 84.4 & 51.4 & 82.3\\
    3& VPT & 75.5 & 95.7 & 56.7 & 90.3 & 45.6 & 85.7 & 61.9 & 91.6\\
    4& ImageBind & 59.9 & 83.1 & 25.2 & 49.8 & 42.4 & 73.8 & 30.7 & 56.8\\
    5& LanguageBind & 81.0 & 97.6 & 47.3 & 81.2 & 58.5 & 87.9 & 55.5 & 85.6\\
    \midrule
    \rowcolor{aliceblue!60} 6& FreestyleRet-CLIP & 80.2 & 97.5 & 62.6 & \textbf{95.2} & 68.7 & 96.6 & 67.4 & \textbf{95.3}\\
    \rowcolor{aliceblue!60} 7& FreestyleRet-BLIP & 88.4 & \textbf{98.6} & \textbf{63.9} & 94.1 & \textbf{76.0} & \textbf{97.5} & \textbf{71.3} & 94.3\\

 \bottomrule[1.5pt]
\end{tabular}
}
}
\caption{
\textbf{The Low-Resolution Image Retrieval performance of our FreestyleRet and baseline models.} 
}
\label{tab:result_Low_Res}
\end{table*}

\section{Comparison with Other Retrieval Settings}
Our FreestyleRet proposes a novel retrieval setting: Image Retrieval with Style-Diversified Queries.  However, during our survey of related works, we have identified several closely related retrieval tasks, including Composed Image Retrieval~\cite{vo2019composing}, User Generalized Image Retrieval~\cite{ma2021model}, Fashion Retrieval~\cite{han2017automatic}, Synthesis Image Retrieval~\cite{vo2019composing}, and Sketch Retrieval~\cite{xu2018sketchmate}. Consequently, we summarize these tasks and highlight the differences and contributions of our novel task: Style-Diversified Image Retrieval Task in comparison to them.

\subsection{Composed Image Retrieval}
\noindent \textbf{Introduction:} Composed Image Retrieval~(CIR)~\cite{liu2021image,vo2019composing} aims to retrieve a target image based on a query composed of a reference image and a relative caption that describes the difference between the two images. Zero-shot CIR~\cite{baldrati2023zero} is a derivative task associated with CIR, learning image-text joint features without requiring a labeled training dataset. The CIR task has been extensively studied in various Vision and Language tasks, such as visual question answering~\cite{li2023tg,ye2023fits} and visual grounding~\cite{cheng2023wico,cheng2023parallel}.

\hspace*{\fill}

\noindent \textbf{Difference:} 
The composed image retrieval focuses on retrieving natural images from composed queries~(image+text) and does not consider style-diversified query inputs. However, our style-diversified retrieval setting not only achieves style-diversified query-based retrieval ability but also achieves good performance when retrieving from composed queries with various styles~(sketch+text, art+text, low resolution+text).

\subsection{User Generalized Image Retrieval}
\noindent \textbf{Introduction:} The User Generalized Image Retrieval (UGIR)~\cite{ma2021model} is a task that retrieves natural images and text. Formally, UGIR defines data belonging to one user as a user domain, and the differences among different user domains as user domain shift. UGIR trains on a user domain and tests on various user domains to evaluate their feature generalization.

\hspace*{\fill}

\noindent \textbf{Difference:} 
The user-generalized image retrieval task focuses on exploring the domain adaptation capability of retrieval models, where the domain refers to a natural image dataset encompassing diverse categories of objects. 
However, in our style-diversified retrieval setting, we adapt the domain of a wide range of image styles as queries, including natural images, sketches, artistic images, and blurry low-resolution images.

\subsection{Fashion, Synthesis, and Sketch Retrieval}
\noindent \textbf{Introduction:} Fashion Retrieval~\cite{han2017automatic,Guo2018,sui2023dtrn}, Synthesis Image Retrieval~\cite{vo2019composing,Zhao2018,sui2023gcrdn}, and Sketch Retrieval~\cite{sangkloy2022sketch} aim to retrieve from one specific class of images, including the fashion clothes, synthesis natural scenes, and sketch-based images. These tasks are applied in the search engines.

\hspace*{\fill}

\noindent \textbf{Difference:} 
The fashion retrieval, synthesis retrieval, and sketch retrieval all focus on retrieving from single-style queries. However, our style-diversified retrieval maintains the ability to retrieve based on queries with various styles, including sketch images and synthesis art-style images.

\section{Supplements for Case Study}

\begin{figure*}
    \centering
    \includegraphics[width=1.0\textwidth]{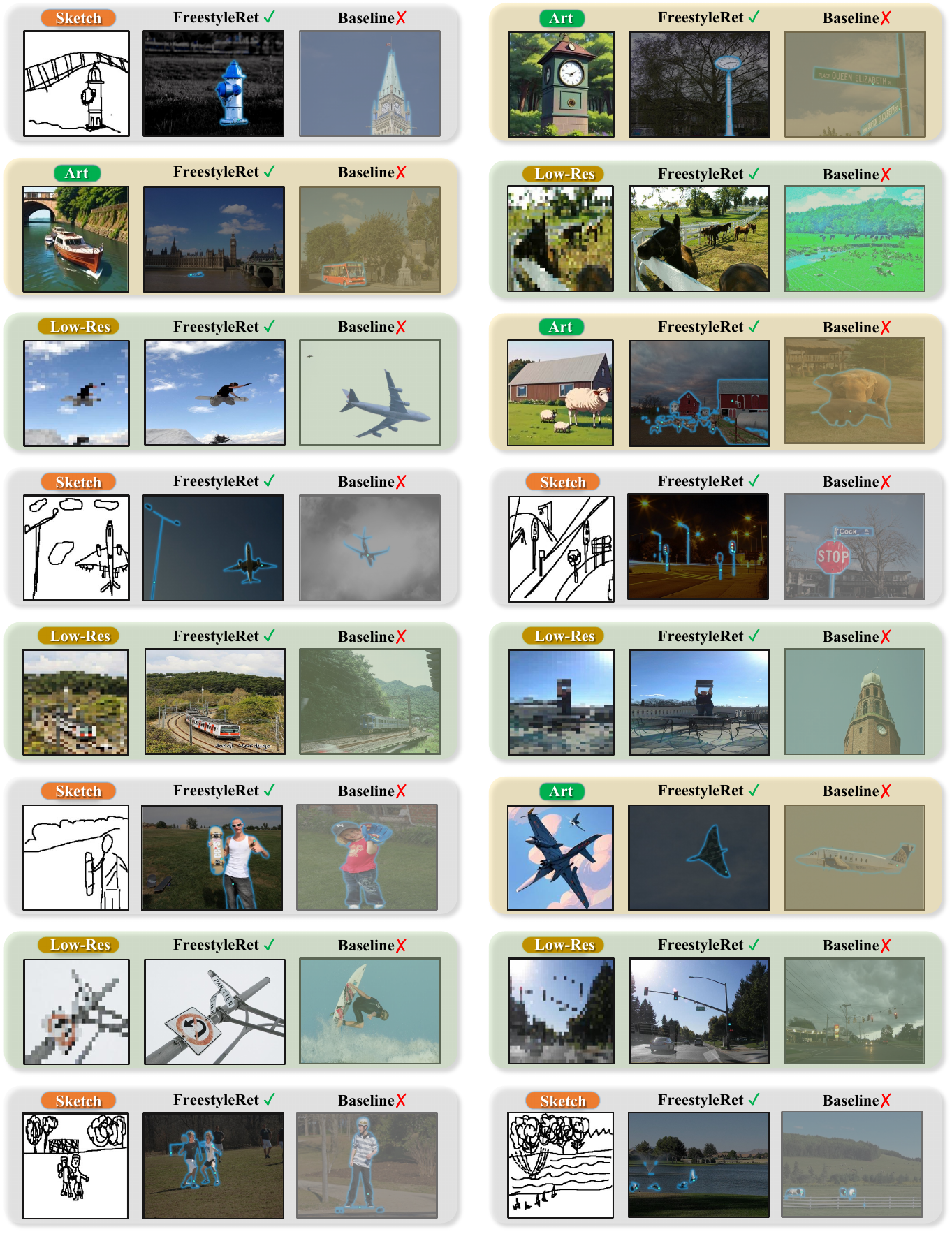}
    \caption{The Visualization of our FreestyleRet-BLIP and the baseline BLIP model on our DSR dataset.}
    \label{fig:case1}
\end{figure*}

\begin{figure*}
    \centering
    \includegraphics[width=1.0\textwidth]{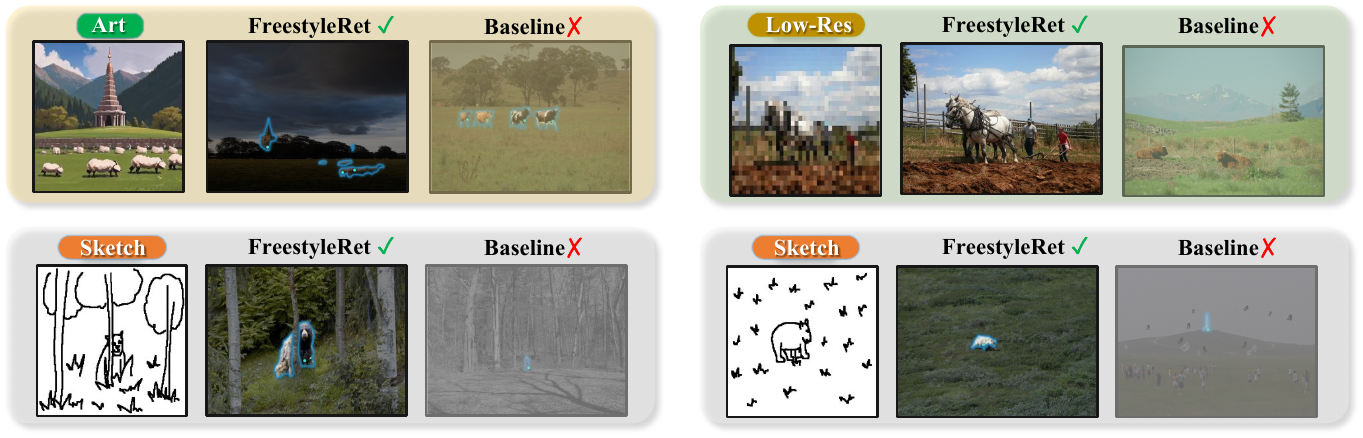}
    \caption{The Visualization of our FreestyleRet-BLIP and the baseline BLIP model on our DSR dataset.}
    \label{fig:case2}
\end{figure*}

As shown in Fig.~\ref{fig:case1} and Fig.~\ref{fig:case2}, we add more visualization results in our supplementary material. Each sample has three images to compare the retrieval performance between our FreestyleRet and the CLIP baseline on the DSR dataset. The left images are the queries randomly selected from different styles. The middle and the right images are the retrieval results of our FreestyleRet-BLIP model and the original BLIP model, respectively.

%% file: main.bbl
\begin{thebibliography}{63}
\providecommand{\natexlab}[1]{#1}
\providecommand{\url}[1]{\texttt{#1}}
\expandafter\ifx\csname urlstyle\endcsname\relax
  \providecommand{\doi}[1]{doi: #1}\else
  \providecommand{\doi}{doi: \begingroup \urlstyle{rm}\Url}\fi

\bibitem[Baldrati et~al.(2023)Baldrati, Agnolucci, Bertini, and Del~Bimbo]{baldrati2023zero}
Alberto Baldrati, Lorenzo Agnolucci, Marco Bertini, and Alberto Del~Bimbo.
\newblock Zero-shot composed image retrieval with textual inversion.
\newblock \emph{arXiv preprint arXiv:2303.15247}, 2023.

\bibitem[Bossett et~al.(2021)Bossett, Heimowitz, Jadhav, Johnson, Singh, Zheng, and Dasgupta]{bossett2021emotion}
Daniel Bossett, David Heimowitz, Nidhi Jadhav, Leilani Johnson, Arti Singh, Helen Zheng, and Sabar Dasgupta.
\newblock Emotion-based style transfer on visual art using gram matrices.
\newblock In \emph{2021 IEEE MIT Undergraduate Research Technology Conference (URTC)}, pages 1--5. IEEE, 2021.

\bibitem[Brown et~al.(2020)Brown, Mann, Ryder, Subbiah, Kaplan, Dhariwal, Neelakantan, Shyam, Sastry, Askell, et~al.]{brown2020language}
Tom Brown, Benjamin Mann, Nick Ryder, Melanie Subbiah, Jared~D Kaplan, Prafulla Dhariwal, Arvind Neelakantan, Pranav Shyam, Girish Sastry, Amanda Askell, et~al.
\newblock Language models are few-shot learners.
\newblock \emph{Advances in neural information processing systems}, 33:\penalty0 1877--1901, 2020.

\bibitem[Cai et~al.(2023)Cai, Ma, Wang, and Li]{cai2023image}
Qiang Cai, Mengxu Ma, Chen Wang, and Haisheng Li.
\newblock Image neural style transfer: A review.
\newblock \emph{Computers and Electrical Engineering}, 108:\penalty0 108723, 2023.

\bibitem[Cheng et~al.()Cheng, Zhang, Yu, Wang, Li, and Zhang]{chengnull}
Xinhua Cheng, Nan Zhang, Jiwen Yu, Yinhuai Wang, Ge Li, and Jian Zhang.
\newblock Null-space diffusion sampling for zero-shot point cloud completion.

\bibitem[Cheng et~al.(2023{\natexlab{a}})Cheng, Jin, Li, Li, Li, Ji, Liu, and Chen]{cheng2023wico}
Zesen Cheng, Peng Jin, Hao Li, Kehan Li, Siheng Li, Xiangyang Ji, Chang Liu, and Jie Chen.
\newblock Wico: Win-win cooperation of bottom-up and top-down referring image segmentation.
\newblock \emph{arXiv preprint arXiv:2306.10750}, 2023{\natexlab{a}}.

\bibitem[Cheng et~al.(2023{\natexlab{b}})Cheng, Li, Jin, Ji, Yuan, Liu, and Chen]{cheng2023parallel}
Zesen Cheng, Kehan Li, Peng Jin, Xiangyang Ji, Li Yuan, Chang Liu, and Jie Chen.
\newblock Parallel vertex diffusion for unified visual grounding.
\newblock \emph{arXiv preprint arXiv:2303.07216}, 2023{\natexlab{b}}.

\bibitem[Chowdhury et~al.(2022)Chowdhury, Sain, Bhunia, Xiang, Gryaditskaya, and Song]{chowdhury2022fs}
Pinaki~Nath Chowdhury, Aneeshan Sain, Ayan~Kumar Bhunia, Tao Xiang, Yulia Gryaditskaya, and Yi-Zhe Song.
\newblock Fs-coco: Towards understanding of freehand sketches of common objects in context.
\newblock In \emph{European Conference on Computer Vision}, pages 253--270. Springer, 2022.

\bibitem[Chowdhury et~al.(2023)Chowdhury, Bhunia, Sain, Koley, Xiang, and Song]{chowdhury2023scenetrilogy}
Pinaki~Nath Chowdhury, Ayan~Kumar Bhunia, Aneeshan Sain, Subhadeep Koley, Tao Xiang, and Yi-Zhe Song.
\newblock Scenetrilogy: On human scene-sketch and its complementarity with photo and text.
\newblock In \emph{Proceedings of the IEEE/CVF Conference on Computer Vision and Pattern Recognition}, pages 10972--10983, 2023.

\bibitem[Datta et~al.(2008)Datta, Joshi, Li, and Wang]{datta2008image}
Ritendra Datta, Dhiraj Joshi, Jia Li, and James~Z Wang.
\newblock Image retrieval: Ideas, influences, and trends of the new age.
\newblock \emph{ACM Computing Surveys (Csur)}, 40\penalty0 (2):\penalty0 1--60, 2008.

\bibitem[Deng et~al.(2009)Deng, Dong, Socher, Li, Li, and Fei-Fei]{deng2009imagenet}
Jia Deng, Wei Dong, Richard Socher, Li-Jia Li, Kai Li, and Li Fei-Fei.
\newblock Imagenet: A large-scale hierarchical image database.
\newblock In \emph{2009 IEEE conference on computer vision and pattern recognition}, pages 248--255. Ieee, 2009.

\bibitem[Dosovitskiy et~al.(2020)Dosovitskiy, Beyer, Kolesnikov, Weissenborn, Zhai, Unterthiner, Dehghani, Minderer, Heigold, Gelly, et~al.]{dosovitskiy2020image}
Alexey Dosovitskiy, Lucas Beyer, Alexander Kolesnikov, Dirk Weissenborn, Xiaohua Zhai, Thomas Unterthiner, Mostafa Dehghani, Matthias Minderer, Georg Heigold, Sylvain Gelly, et~al.
\newblock An image is worth 16x16 words: Transformers for image recognition at scale.
\newblock \emph{arXiv preprint arXiv:2010.11929}, 2020.

\bibitem[Efros and Freeman(2023)]{efros2023image}
Alexei~A Efros and William~T Freeman.
\newblock Image quilting for texture synthesis and transfer.
\newblock In \emph{Seminal Graphics Papers: Pushing the Boundaries, Volume 2}, pages 571--576. 2023.

\bibitem[Efros and Leung(1999)]{efros1999texture}
Alexei~A Efros and Thomas~K Leung.
\newblock Texture synthesis by non-parametric sampling.
\newblock In \emph{Proceedings of the seventh IEEE international conference on computer vision}, pages 1033--1038. IEEE, 1999.

\bibitem[Gatys et~al.(2016)Gatys, Ecker, and Bethge]{gatys2016image}
Leon~A Gatys, Alexander~S Ecker, and Matthias Bethge.
\newblock Image style transfer using convolutional neural networks.
\newblock In \emph{Proceedings of the IEEE conference on computer vision and pattern recognition}, pages 2414--2423, 2016.

\bibitem[Girdhar et~al.(2023)Girdhar, El-Nouby, Liu, Singh, Alwala, Joulin, and Misra]{girdhar2023imagebind}
Rohit Girdhar, Alaaeldin El-Nouby, Zhuang Liu, Mannat Singh, Kalyan~Vasudev Alwala, Armand Joulin, and Ishan Misra.
\newblock Imagebind: One embedding space to bind them all.
\newblock In \emph{Proceedings of the IEEE/CVF Conference on Computer Vision and Pattern Recognition}, pages 15180--15190, 2023.

\bibitem[Guo et~al.(2018)Guo, Wu, Cheng, Rennie, Tesauro, and Feris]{Guo2018}
Xiaoxiao Guo, Hui Wu, Yu Cheng, StevenJ. Rennie, Gerald Tesauro, and Rogerio Feris.
\newblock Dialog-based interactive image retrieval.
\newblock \emph{arXiv: Computer Vision and Pattern Recognition,arXiv: Computer Vision and Pattern Recognition}, 2018.

\bibitem[Guo et~al.(2023)Guo, Yang, Rao, Wang, Qiao, Lin, and Dai]{guo2023animatediff}
Yuwei Guo, Ceyuan Yang, Anyi Rao, Yaohui Wang, Yu Qiao, Dahua Lin, and Bo Dai.
\newblock Animatediff: Animate your personalized text-to-image diffusion models without specific tuning.
\newblock \emph{arXiv preprint arXiv:2307.04725}, 2023.

\bibitem[Han et~al.(2017)Han, Wu, Huang, Zhang, Zhu, Li, Zhao, and Davis]{han2017automatic}
Xintong Han, Zuxuan Wu, Phoenix~X Huang, Xiao Zhang, Menglong Zhu, Yuan Li, Yang Zhao, and Larry~S Davis.
\newblock Automatic spatially-aware fashion concept discovery.
\newblock In \emph{Proceedings of the IEEE international conference on computer vision}, pages 1463--1471, 2017.

\bibitem[He et~al.(2016)He, Zhang, Ren, and Sun]{He7780459}
Kaiming He, Xiangyu Zhang, Shaoqing Ren, and Jian Sun.
\newblock Deep residual learning for image recognition.
\newblock In \emph{2016 IEEE Conference on Computer Vision and Pattern Recognition (CVPR)}, pages 770--778, 2016.

\bibitem[Isinkaye et~al.(2015)Isinkaye, Folajimi, and Ojokoh]{isinkaye2015recommendation}
Folasade~Olubusola Isinkaye, Yetunde~O Folajimi, and Bolande~Adefowoke Ojokoh.
\newblock Recommendation systems: Principles, methods and evaluation.
\newblock \emph{Egyptian informatics journal}, 16\penalty0 (3):\penalty0 261--273, 2015.

\bibitem[Jia et~al.(2022)Jia, Tang, Chen, Cardie, Belongie, Hariharan, and Lim]{jia2022vpt}
Menglin Jia, Luming Tang, Bor-Chun Chen, Claire Cardie, Serge Belongie, Bharath Hariharan, and Ser-Nam Lim.
\newblock Visual prompt tuning.
\newblock In \emph{European Conference on Computer Vision (ECCV)}, 2022.

\bibitem[Jin et~al.(2023{\natexlab{a}})Jin, Li, Cheng, Huang, Wang, Yuan, Liu, and Chen]{jin2023text}
Peng Jin, Hao Li, Zesen Cheng, Jinfa Huang, Zhennan Wang, Li Yuan, Chang Liu, and Jie Chen.
\newblock Text-video retrieval with disentangled conceptualization and set-to-set alignment.
\newblock \emph{arXiv preprint arXiv:2305.12218}, 2023{\natexlab{a}}.

\bibitem[Jin et~al.(2023{\natexlab{b}})Jin, Li, Cheng, Li, Ji, Liu, Yuan, and Chen]{jin2023diffusionret}
Peng Jin, Hao Li, Zesen Cheng, Kehan Li, Xiangyang Ji, Chang Liu, Li Yuan, and Jie Chen.
\newblock Diffusionret: Generative text-video retrieval with diffusion model.
\newblock \emph{arXiv preprint arXiv:2303.09867}, 2023{\natexlab{b}}.

\bibitem[Johnson et~al.(2015)Johnson, Krishna, Stark, Li, Shamma, Bernstein, and Fei-Fei]{johnson2015image}
Justin Johnson, Ranjay Krishna, Michael Stark, Li-Jia Li, David Shamma, Michael Bernstein, and Li Fei-Fei.
\newblock Image retrieval using scene graphs.
\newblock In \emph{Proceedings of the IEEE conference on computer vision and pattern recognition}, pages 3668--3678, 2015.

\bibitem[Kafai et~al.(2014)Kafai, Eshghi, and Bhanu]{kafai2014discrete}
Mehran Kafai, Kave Eshghi, and Bir Bhanu.
\newblock Discrete cosine transform locality-sensitive hashes for face retrieval.
\newblock \emph{IEEE Transactions on multimedia}, 16\penalty0 (4):\penalty0 1090--1103, 2014.

\bibitem[Karras et~al.(2019)Karras, Laine, and Aila]{karras2019style}
Tero Karras, Samuli Laine, and Timo Aila.
\newblock A style-based generator architecture for generative adversarial networks.
\newblock In \emph{Proceedings of the IEEE/CVF conference on computer vision and pattern recognition}, pages 4401--4410, 2019.

\bibitem[Lee et~al.(2010)Lee, Seo, Ryoo, and Yoon]{lee2010directional}
Hochang Lee, Sanghyun Seo, Seungtaek Ryoo, and Kyunghyun Yoon.
\newblock Directional texture transfer.
\newblock In \emph{Proceedings of the 8th International Symposium on Non-Photorealistic Animation and Rendering}, pages 43--48, 2010.

\bibitem[Lester et~al.(2021)Lester, Al-Rfou, and Constant]{lester2021power}
Brian Lester, Rami Al-Rfou, and Noah Constant.
\newblock The power of scale for parameter-efficient prompt tuning.
\newblock \emph{arXiv preprint arXiv:2104.08691}, 2021.

\bibitem[Li et~al.(2022{\natexlab{a}})Li, Li, Karimi, Chen, and Sun]{li2022joint}
Hao Li, Xu Li, Belhal Karimi, Jie Chen, and Mingming Sun.
\newblock Joint learning of object graph and relation graph for visual question answering.
\newblock In \emph{2022 IEEE International Conference on Multimedia and Expo (ICME)}, pages 01--06. IEEE, 2022{\natexlab{a}}.

\bibitem[Li et~al.(2023{\natexlab{a}})Li, Huang, Jin, Song, Wu, and Chen]{li2023weakly}
Hao Li, Jinfa Huang, Peng Jin, Guoli Song, Qi Wu, and Jie Chen.
\newblock Weakly-supervised 3d spatial reasoning for text-based visual question answering.
\newblock \emph{IEEE Transactions on Image Processing}, 2023{\natexlab{a}}.

\bibitem[Li et~al.(2023{\natexlab{b}})Li, Jin, Cheng, Zhang, Chen, Wang, Liu, and Chen]{li2023tg}
Hao Li, Peng Jin, Zesen Cheng, Songyang Zhang, Kai Chen, Zhennan Wang, Chang Liu, and Jie Chen.
\newblock Tg-vqa: Ternary game of video question answering.
\newblock \emph{arXiv preprint arXiv:2305.10049}, 2023{\natexlab{b}}.

\bibitem[Li et~al.(2022{\natexlab{b}})Li, Li, Xiong, and Hoi]{li2022blip}
Junnan Li, Dongxu Li, Caiming Xiong, and Steven Hoi.
\newblock Blip: Bootstrapping language-image pre-training for unified vision-language understanding and generation.
\newblock In \emph{International Conference on Machine Learning}, pages 12888--12900. PMLR, 2022{\natexlab{b}}.

\bibitem[Li et~al.(2021)Li, Yang, and Ma]{li2021recent}
Xiaoqing Li, Jiansheng Yang, and Jinwen Ma.
\newblock Recent developments of content-based image retrieval (cbir).
\newblock \emph{Neurocomputing}, 452:\penalty0 675--689, 2021.

\bibitem[Li and Liang(2021)]{li2021prefix}
Xiang~Lisa Li and Percy Liang.
\newblock Prefix-tuning: Optimizing continuous prompts for generation.
\newblock \emph{arXiv preprint arXiv:2101.00190}, 2021.

\bibitem[Li et~al.(2017)Li, Fang, Yang, Wang, Lu, and Yang]{li2017universal}
Yijun Li, Chen Fang, Jimei Yang, Zhaowen Wang, Xin Lu, and Ming-Hsuan Yang.
\newblock Universal style transfer via feature transforms.
\newblock \emph{Advances in neural information processing systems}, 30, 2017.

\bibitem[Liu et~al.(2019)Liu, Xi, Ji, and Ma]{LIU2019465}
Long Liu, Zhixuan Xi, RuiRui Ji, and Weigang Ma.
\newblock Advanced deep learning techniques for image style transfer: A survey.
\newblock \emph{Signal Processing: Image Communication}, 78:\penalty0 465--470, 2019.

\bibitem[Liu et~al.(2021{\natexlab{a}})Liu, Ji, Fu, Tam, Du, Yang, and Tang]{liu2021p}
Xiao Liu, Kaixuan Ji, Yicheng Fu, Weng~Lam Tam, Zhengxiao Du, Zhilin Yang, and Jie Tang.
\newblock P-tuning v2: Prompt tuning can be comparable to fine-tuning universally across scales and tasks.
\newblock \emph{arXiv preprint arXiv:2110.07602}, 2021{\natexlab{a}}.

\bibitem[Liu et~al.(2022)Liu, Ji, Fu, Tam, Du, Yang, and Tang]{liu2022p}
Xiao Liu, Kaixuan Ji, Yicheng Fu, Weng Tam, Zhengxiao Du, Zhilin Yang, and Jie Tang.
\newblock P-tuning: Prompt tuning can be comparable to fine-tuning across scales and tasks.
\newblock In \emph{Proceedings of the 60th Annual Meeting of the Association for Computational Linguistics (Volume 2: Short Papers)}, pages 61--68, 2022.

\bibitem[Liu et~al.(2021{\natexlab{b}})Liu, Rodriguez-Opazo, Teney, and Gould]{liu2021image}
Zheyuan Liu, Cristian Rodriguez-Opazo, Damien Teney, and Stephen Gould.
\newblock Image retrieval on real-life images with pre-trained vision-and-language models.
\newblock In \emph{Proceedings of the IEEE/CVF International Conference on Computer Vision}, pages 2125--2134, 2021{\natexlab{b}}.

\bibitem[Ma et~al.(2021)Ma, Yang, Gao, and Xu]{ma2021model}
Xinhong Ma, Xiaoshan Yang, Junyu Gao, and Changsheng Xu.
\newblock The model may fit you: User-generalized cross-modal retrieval.
\newblock \emph{IEEE Transactions on Multimedia}, 24:\penalty0 2998--3012, 2021.

\bibitem[Radford et~al.(2021)Radford, Kim, Hallacy, Ramesh, Goh, Agarwal, Sastry, Askell, Mishkin, Clark, et~al.]{radford2021learning}
Alec Radford, Jong~Wook Kim, Chris Hallacy, Aditya Ramesh, Gabriel Goh, Sandhini Agarwal, Girish Sastry, Amanda Askell, Pamela Mishkin, Jack Clark, et~al.
\newblock Learning transferable visual models from natural language supervision.
\newblock In \emph{International conference on machine learning}, pages 8748--8763. PMLR, 2021.

\bibitem[Richardson et~al.(2021)Richardson, Alaluf, Patashnik, Nitzan, Azar, Shapiro, and Cohen-Or]{richardson2021encoding}
Elad Richardson, Yuval Alaluf, Or Patashnik, Yotam Nitzan, Yaniv Azar, Stav Shapiro, and Daniel Cohen-Or.
\newblock Encoding in style: a stylegan encoder for image-to-image translation.
\newblock In \emph{Proceedings of the IEEE/CVF conference on computer vision and pattern recognition}, pages 2287--2296, 2021.

\bibitem[Sangkloy et~al.(2022)Sangkloy, Jitkrittum, Yang, and Hays]{sangkloy2022sketch}
Patsorn Sangkloy, Wittawat Jitkrittum, Diyi Yang, and James Hays.
\newblock A sketch is worth a thousand words: Image retrieval with text and sketch.
\newblock In \emph{European Conference on Computer Vision}, pages 251--267. Springer, 2022.

\bibitem[Simonyan and Zisserman(2014)]{simonyan2014very}
Karen Simonyan and Andrew Zisserman.
\newblock Very deep convolutional networks for large-scale image recognition.
\newblock \emph{arXiv preprint arXiv:1409.1556}, 2014.

\bibitem[Su et~al.(2021)Su, Liu, Yu, Hu, Liao, Tian, Pietik{\"a}inen, and Liu]{su2021pixel}
Zhuo Su, Wenzhe Liu, Zitong Yu, Dewen Hu, Qing Liao, Qi Tian, Matti Pietik{\"a}inen, and Li Liu.
\newblock Pixel difference networks for efficient edge detection.
\newblock In \emph{Proceedings of the IEEE/CVF international conference on computer vision}, pages 5117--5127, 2021.

\bibitem[Sui et~al.(2023{\natexlab{a}})Sui, Ma, Zhang, and Pun]{sui2023dtrn}
Jialu Sui, Xianping Ma, Xiaokang Zhang, and Man-On Pun.
\newblock Dtrn: Dual transformer residual network for remote sensing super-resolution.
\newblock In \emph{IGARSS 2023-2023 IEEE International Geoscience and Remote Sensing Symposium}, pages 6041--6044. IEEE, 2023{\natexlab{a}}.

\bibitem[Sui et~al.(2023{\natexlab{b}})Sui, Ma, Zhang, and Pun]{sui2023gcrdn}
Jialu Sui, Xianping Ma, Xiaokang Zhang, and Man-On Pun.
\newblock Gcrdn: Global context-driven residual dense network for remote sensing image super-resolution.
\newblock \emph{IEEE Journal of Selected Topics in Applied Earth Observations and Remote Sensing}, 2023{\natexlab{b}}.

\bibitem[Tao(2022)]{tao2022image}
Yilin Tao.
\newblock Image style transfer based on vgg neural network model.
\newblock In \emph{2022 IEEE International Conference on Advances in Electrical Engineering and Computer Applications (AEECA)}, pages 1475--1482. IEEE, 2022.

\bibitem[Thomee and Lew(2012)]{thomee2012interactive}
Bart Thomee and Michael~S Lew.
\newblock Interactive search in image retrieval: a survey.
\newblock \emph{International Journal of Multimedia Information Retrieval}, 1:\penalty0 71--86, 2012.

\bibitem[Van Der~Maaten(2009)]{van2009learning}
Laurens Van Der~Maaten.
\newblock Learning a parametric embedding by preserving local structure.
\newblock In \emph{Artificial intelligence and statistics}, pages 384--391. PMLR, 2009.

\bibitem[Vo et~al.(2019)Vo, Jiang, Sun, Murphy, Li, Fei-Fei, and Hays]{vo2019composing}
Nam Vo, Lu Jiang, Chen Sun, Kevin Murphy, Li-Jia Li, Li Fei-Fei, and James Hays.
\newblock Composing text and image for image retrieval-an empirical odyssey.
\newblock In \emph{Proceedings of the IEEE/CVF conference on computer vision and pattern recognition}, pages 6439--6448, 2019.

\bibitem[Wang et~al.(2021)Wang, Li, and Vasconcelos]{wang2021rethinking}
Pei Wang, Yijun Li, and Nuno Vasconcelos.
\newblock Rethinking and improving the robustness of image style transfer.
\newblock In \emph{Proceedings of the IEEE/CVF conference on computer vision and pattern recognition}, pages 124--133, 2021.

\bibitem[Wang et~al.(2022)Wang, Yu, and Zhang]{wang2022zero}
Yinhuai Wang, Jiwen Yu, and Jian Zhang.
\newblock Zero-shot image restoration using denoising diffusion null-space model.
\newblock \emph{arXiv preprint arXiv:2212.00490}, 2022.

\bibitem[Xu et~al.(2018)Xu, Huang, Yuan, Pang, Song, Xiang, Hospedales, Ma, and Guo]{xu2018sketchmate}
Peng Xu, Yongye Huang, Tongtong Yuan, Kaiyue Pang, Yi-Zhe Song, Tao Xiang, Timothy~M Hospedales, Zhanyu Ma, and Jun Guo.
\newblock Sketchmate: Deep hashing for million-scale human sketch retrieval.
\newblock In \emph{Proceedings of the IEEE conference on computer vision and pattern recognition}, pages 8090--8098, 2018.

\bibitem[Ye et~al.(2023)Ye, Cao, Chen, Xu, and Zou]{ye2023fits}
Qichen Ye, Bowen Cao, Nuo Chen, Weiyuan Xu, and Yuexian Zou.
\newblock Fits: Fine-grained two-stage training for knowledge-aware question answering.
\newblock \emph{arXiv preprint arXiv:2302.11799}, 2023.

\bibitem[Yu et~al.(2023)Yu, Wang, Zhao, Ghanem, and Zhang]{yu2023freedom}
Jiwen Yu, Yinhuai Wang, Chen Zhao, Bernard Ghanem, and Jian Zhang.
\newblock Freedom: Training-free energy-guided conditional diffusion model.
\newblock \emph{arXiv preprint arXiv:2303.09833}, 2023.

\bibitem[Zha et~al.(2023)Zha, Wang, Dai, Chen, Wang, and Xia]{zha2023instance}
Yaohua Zha, Jinpeng Wang, Tao Dai, Bin Chen, Zhi Wang, and Shu-Tao Xia.
\newblock Instance-aware dynamic prompt tuning for pre-trained point cloud models.
\newblock \emph{arXiv preprint arXiv:2304.07221}, 2023.

\bibitem[Zhao(2020)]{zhao2020survey}
Changshen Zhao.
\newblock A survey on image style transfer approaches using deep learning.
\newblock In \emph{Journal of Physics: Conference Series}, page 012129. IOP Publishing, 2020.

\bibitem[Zhao et~al.(2018)Zhao, Wang, and Qi]{Zhao2018}
Yu Zhao, Jingyu Wang, and Qi Qi.
\newblock Mindcamera: Interactive image retrieval and synthesis.
\newblock In \emph{2018 IEEE 13th Image, Video, and Multidimensional Signal Processing Workshop (IVMSP)}, 2018.

\bibitem[Zhou et~al.(2022{\natexlab{a}})Zhou, Yang, Loy, and Liu]{zhou2022conditional}
Kaiyang Zhou, Jingkang Yang, Chen~Change Loy, and Ziwei Liu.
\newblock Conditional prompt learning for vision-language models.
\newblock In \emph{Proceedings of the IEEE/CVF Conference on Computer Vision and Pattern Recognition}, pages 16816--16825, 2022{\natexlab{a}}.

\bibitem[Zhou et~al.(2022{\natexlab{b}})Zhou, Yang, Loy, and Liu]{zhou2022learning}
Kaiyang Zhou, Jingkang Yang, Chen~Change Loy, and Ziwei Liu.
\newblock Learning to prompt for vision-language models.
\newblock \emph{International Journal of Computer Vision}, 130\penalty0 (9):\penalty0 2337--2348, 2022{\natexlab{b}}.

\bibitem[Zhu et~al.(2023)Zhu, Lin, Ning, Yan, Cui, Wang, Pang, Jiang, Zhang, Li, Zhang, Li, Liu, and Yuan]{Zhu2023LanguageBindEV}
Bin Zhu, Bin Lin, Munan Ning, Yang Yan, Jiaxi Cui, HongFa Wang, Yatian Pang, Wenhao Jiang, Junwu Zhang, Zongwei Li, Wancai Zhang, Zhifeng Li, Wei Liu, and Liejie Yuan.
\newblock Languagebind: Extending video-language pretraining to n-modality by language-based semantic alignment.
\newblock \emph{ArXiv}, abs/2310.01852, 2023.

\end{thebibliography}
